\documentclass[11pt]{article}

\usepackage[preprint]{acl}

\usepackage{times}
\usepackage{latexsym}
\usepackage{url}

\usepackage[T1]{fontenc}

\usepackage[utf8]{inputenc}

\usepackage{microtype}
\usepackage[english]{babel}
\usepackage{csquotes}
\usepackage{inconsolata}

\usepackage{graphicx}
\usepackage{booktabs}
\usepackage{enumitem}
\usepackage{multirow}
\usepackage{tabularx}
\usepackage{hyperref}

\usepackage{subcaption}
\usepackage{adjustbox}
\usepackage{placeins}

\usepackage{array}
\usepackage{tcolorbox}

\usepackage{tikz}
\usetikzlibrary{arrows.meta, positioning, shapes.geometric}

\usepackage{amsmath}

\usepackage{pgfplots}
\pgfplotsset{compat=1.17}


\newcommand\dataset{\texttt{ClaimFlow}}
\newcommand\datasetauto{\texttt{ClaimFlow-\allowbreak AutoGraph}}
\newcommand{\ukpauthorlogo}{\raisebox{0.65ex}{\includegraphics[height=0.75em]{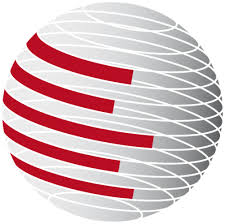}}}
\newcommand{\ituauthorlogo}{\raisebox{0.78ex}{\includegraphics[width=1.45em,trim=12 62 12 62,clip]{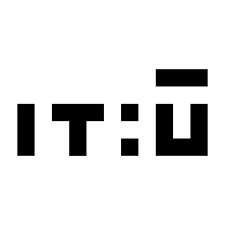}}}
\newcommand{\nrcauthorlogo}{\raisebox{0.82ex}{\includegraphics[width=2.35em,trim=0 84 0 84,clip]{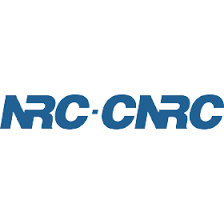}}}
\newcommand{\ukpaffillogo}{\raisebox{0.65ex}{\includegraphics[height=0.85em]{logos/ukp-logo.jpeg}}}
\newcommand{\ituaffillogo}{\raisebox{0.7ex}{\includegraphics[width=1.55em,trim=12 62 12 62,clip]{logos/itu-logo.png}}}
\newcommand{\nrcaffillogo}{\raisebox{0.72ex}{\includegraphics[width=2.25em,trim=0 84 0 84,clip]{logos/nrc-logo.png}}}

\definecolor{claimflowbackground}{HTML}{357EDD}
\definecolor{claimflowsupport}{HTML}{19A974}
\definecolor{claimflowextend}{HTML}{FFB700}
\definecolor{claimflowqualify}{HTML}{5E2CA5}
\definecolor{claimflowrefute}{HTML}{E7040F}
\DeclareRobustCommand{\claimflowlogo}{%
  \tikz[baseline=-0.68ex, x=1.35ex, y=1.35ex, line cap=round, line join=round]{%
    \path[use as bounding box] (-1.25,-0.7) rectangle (1.25,0.72);
    \fill[claimflowbackground] (-0.86,0) circle[radius=0.3];
    \fill[claimflowbackground] (0.86,0) circle[radius=0.3];
    \fill[white] (-0.94,0.07) circle[radius=0.065];
    \fill[white] (-0.81,-0.05) circle[radius=0.065];
    \fill[white] (0.78,0.07) circle[radius=0.065];
    \fill[white] (0.91,-0.05) circle[radius=0.065];
    \draw[-{Latex[length=0.42ex,width=0.34ex]}, claimflowsupport, line width=0.52pt]
      (-0.5,0.24) .. controls (-0.16,0.63) and (0.16,0.63) .. (0.5,0.24);
    \draw[-{Latex[length=0.42ex,width=0.34ex]}, claimflowextend, line width=0.52pt]
      (-0.5,0.08) .. controls (-0.17,0.23) and (0.17,0.23) .. (0.5,0.08);
    \draw[-{Latex[length=0.42ex,width=0.34ex]}, claimflowqualify, line width=0.52pt]
      (-0.5,-0.08) .. controls (-0.17,-0.23) and (0.17,-0.23) .. (0.5,-0.08);
    \draw[-{Latex[length=0.42ex,width=0.34ex]}, claimflowrefute, line width=0.52pt]
      (-0.5,-0.24) .. controls (-0.16,-0.63) and (0.16,-0.63) .. (0.5,-0.24);
  }%
}

\title{\claimflowlogo\hspace{0.15em}\dataset{}: Tracing the Evolution of Scientific Claims in NLP}

\author{
Aniket Pramanick\ukpauthorlogo{},
Yufang Hou\ituauthorlogo{},
Saif M. Mohammad\nrcauthorlogo{},
Iryna Gurevych\ukpauthorlogo{} \\[0.5ex]
\ukpaffillogo{} Ubiquitous Knowledge Processing Lab (UKP Lab) \\
Department of Computer Science and Hessian Center for AI (hessian.AI) \\
Technische Universität Darmstadt \\
\ituaffillogo{} IT:U Interdisciplinary Transformation University Austria \\
\nrcaffillogo{} National Research Council Canada \\[0.5ex]
\small{\url{www.ukp.tu-darmstadt.de},
\url{yufang.hou@it-u.at},
\url{saif.mohammad@nrc-cnrc.gc.ca}}
}

\begin{document}
\maketitle

\begin{abstract}

Scientific papers advance \textit{claims} that later work supports, extends, or sometimes refutes. Yet existing methods for citation and claim analysis capture only fragments of this dialogue. In this work, we make these interactions explicit at the level of individual scientific claims. We introduce \dataset{}, a claim-centric view of the NLP literature, built from $1{,}617$ ACL Anthology papers (1979--2025) that are manually annotated with $5{,}689$ claims and $4{,}871$ cross-paper claim relations, indicating whether a citing paper \textcolor{claimflowsupport}{\texttt{supports}}, \textcolor{claimflowextend}{\texttt{extends}}, \textcolor{claimflowqualify}{\texttt{qualifies}}, \textcolor{claimflowrefute}{\texttt{refutes}}, or references a cited claim as \textcolor{claimflowbackground}{\texttt{background}}. Building on \dataset{}, we define a new task -- \textit{Claim Relation Classification} -- which requires models to infer the scientific stance toward a cited claim from the text and citation context. Evaluating neural models and large language models on this task, we report baseline performance of $0.81$ macro-F1, suggesting that the task is tractable while leaving room for improvement. We then scale this framework to $\sim$$13k$ NLP papers to study claim evolution across decades of NLP research. We show that $63.5\%$ claims are never reused; only $11.1\%$ are ever challenged. Widely propagated claims are more often \textit{reshaped} through qualification and extension than supported or refuted. Overall \dataset{} offers a lens for examining how ideas shift and mature within NLP.~\footnote{Code and data available at \url{https://github.com/UKPLab/arxiv2026-claim-flow}}

\end{abstract}

\section{Introduction}

\begin{figure}
    \centering
    \scalebox{0.80}{
    \includegraphics[width=0.95\columnwidth]{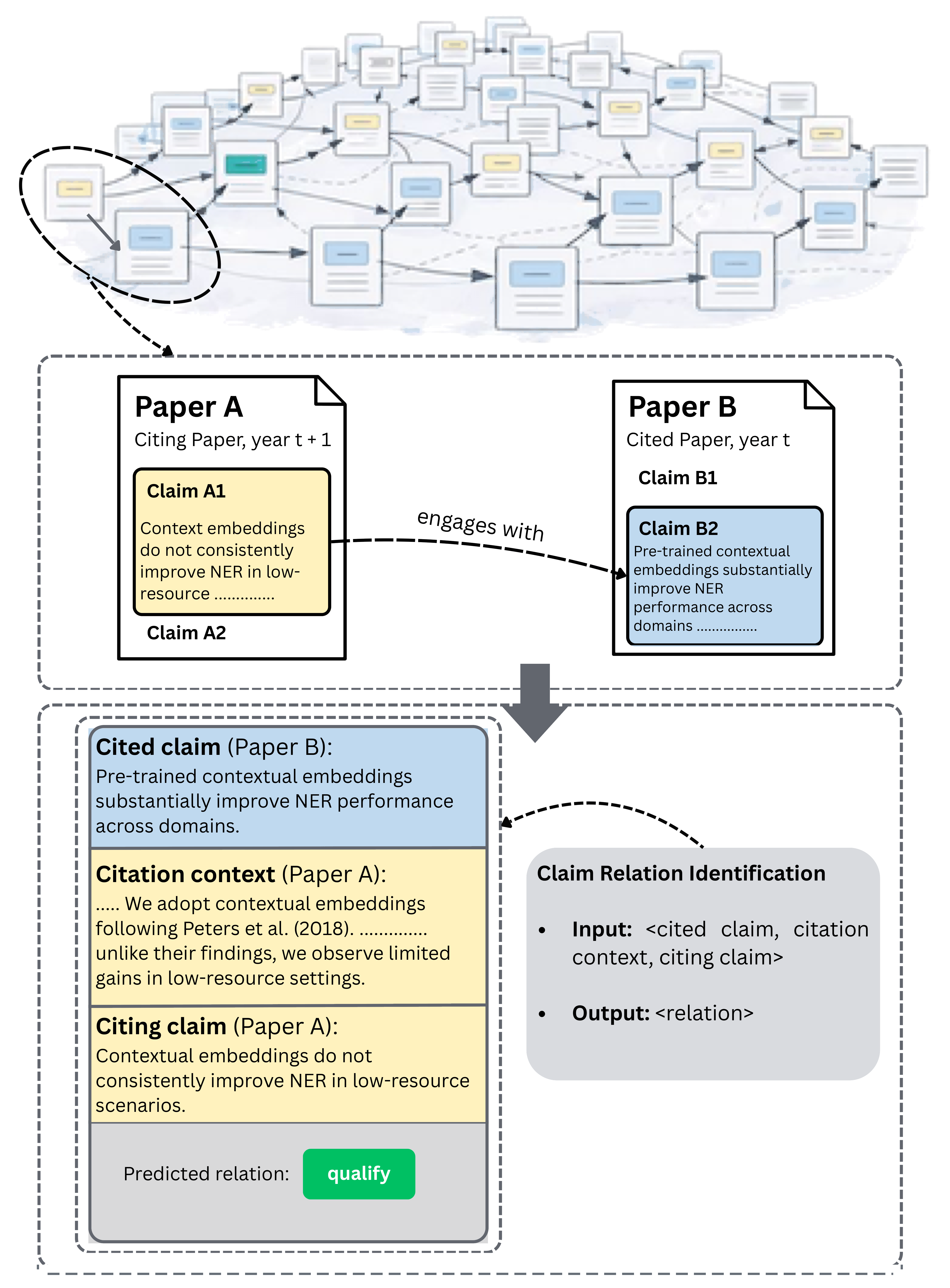}}
    \caption{
    Claims form a directed graph across papers: edges link citing and cited claims, and their relations are predicted from claims and citation context.
    }
    \label{fig:fig1}
\end{figure}

Scientific research is often described as a steady accumulation of knowledge, but what actually accumulates are \textit{claims}; assertions about how the world works, why a method succeeds, or under what conditions an idea fails. These claims are the basic currency of scientific inquiry~\citep{popper1959logic, lakatos2014falsification}.
A new paper often includes claims that reinforce, extend (qualify), or even challenge (overturn) past claims, yet these interactions remain poorly understood.

NLP is a particularly interesting field to study this process. It spans computer science, linguistics, and social science~\citep{manning1999foundations, jurafsky2024speech}; the field has repeatedly refined its central questions~\citep{pramanick-etal-2023-diachronic}.
These shifts are reflected in the kinds of scientific claims researchers make: from early symbolic accounts of parsing~\citep{winograd1972understanding} to statistical alignment models~\citep{brown-etal-1993-mathematics} to deep contextual representations~\citep{devlin-etal-2019-bert}, and more recently to claims about scaling laws and reasoning abilities in LLMs~\citep{wei2022chain}.
Yet we still lack systematic tools for organizing claims into a coherent record of how the field has developed.

Understanding the scientific character of NLP requires moving beyond papers as atomic units towards the claims they advance and how later work engages with them.
Each research paper introduces one or more central claims -- about model behavior, properties of data and language, or assumptions about human language use -- and later citations implicitly position new work relative to these claims. Such engagement may support a claim, extend it to new settings, qualify its scope, or actively refute it.

In this paper, we make claim-level scientific interaction explicit. We introduce \dataset{}, a claim-centric view of the NLP literature constructed from $1{,}617$ ACL Anthology papers from $1979$ to $2025$~\citep{bird2008acl}, manually annotated with $5{,}689$ scientific claims and $4{,}871$ cross-paper claim-level relations.
Relations are labeled as \textcolor{claimflowbackground}{\texttt{background}}, \textcolor{claimflowsupport}{\texttt{support}}, \textcolor{claimflowextend}{\texttt{extension}}, \textcolor{claimflowqualify}{\texttt{qualification}}, or \textcolor{claimflowrefute}{\texttt{refutation}}.
Building on \dataset{}, we define \textit{Claim Relation Classification}, a task that takes as input (i) a \textit{cited claim} from an earlier paper, (ii) a \textit{citing claim} advanced by a later paper, and (iii) the \textit{citation context} -- the sentence containing the citation marker together with its immediately preceding and following sentences -- to determine whether later work adopts, extends, or challenges a prior claim. We build on existing claim identification methods~\citep{pramanick-etal-2025-nature} to focus on modeling how claims in later papers engage with prior work. Figure~\ref{fig:fig1} illustrates the claim graph and the Claim Relation Classification task.

To study whether current models can recover claim-level relations, we develop baseline systems for \textit{Claim Relation Classification}. Models achieve reasonable overall performance (e.g., macro-F1 $0.81$), but struggle with subtle distinctions such as extension versus qualification. We apply the best-performing model to recover claim relations across $13{,}358$ ACL Anthology papers spanning 1979--2025, enabling longitudinal analysis of how scientific ideas evolve within NLP research. Our analyses reveal that challenges are rare and typically short-lived, while widely propagated claims are more often refined through qualification and extension than directly confirmed or rejected.

\paragraph{Contributions.}
Our contributions are threefold: (1) we introduce \dataset{}, an expert-annotated dataset of $1{,}617$ NLP papers that explicitly links $5{,}689$ scientific claims across papers and labels their epistemic relations, enabling fine-grained analysis of how claims are reused, extended, or challenged (\S~\ref{sec:data}); (2) we define \textit{Claim Relation Classification}, a new task that requires models to infer the stance a citing claim takes toward a cited claim given the citation context (\S~\ref{sec:task}); and finally, (3) we leverage this framework to present the first large-scale, claim-level longitudinal analysis of NLP research, examining how scientific claims propagate, are challenged, and evolve across $\sim$$13k$ papers published between 1970 and 2025 (\S~\ref{sec:analysis}).

\section{Related Work}
\label{sec:related_work}

\begin{table*}[t]
    \centering
\begingroup
\definecolor{cfBackgroundBg}{HTML}{EBF3FD}
\definecolor{cfBackgroundFg}{HTML}{357EDD}
\definecolor{cfSupportBg}{HTML}{E8F7F1}
\definecolor{cfSupportFg}{HTML}{19A974}
\definecolor{cfExtendBg}{HTML}{FFF4CC}
\definecolor{cfExtendFg}{HTML}{FFB700}
\definecolor{cfQualifyBg}{HTML}{EFEAF6}
\definecolor{cfQualifyFg}{HTML}{5E2CA5}
\definecolor{cfRefuteBg}{HTML}{FDE6E7}
\definecolor{cfRefuteFg}{HTML}{E7040F}
\newcommand{\cfrel}[3]{\begingroup\setlength{\fboxsep}{1.3pt}\colorbox{#1}{\textcolor{#2}{\scriptsize\textsc{#3}}}\endgroup}
\newcommand{\cfbackground}{\cfrel{cfBackgroundBg}{cfBackgroundFg}{background}}
\newcommand{\cfsupport}{\cfrel{cfSupportBg}{cfSupportFg}{support}}
\newcommand{\cfextend}{\cfrel{cfExtendBg}{cfExtendFg}{extend}}
\newcommand{\cfqualify}{\cfrel{cfQualifyBg}{cfQualifyFg}{qualify}}
\newcommand{\cfrefute}{\cfrel{cfRefuteBg}{cfRefuteFg}{refute}}
    \scalebox{0.90}{
    \begin{adjustbox}{width=2.2\columnwidth, center}
        \begin{tabular}{@{}>{\raggedright\arraybackslash}p{2.6cm} >{\raggedright\arraybackslash}p{8.2cm} >{\raggedright\arraybackslash}p{8.2cm} >{\raggedright\arraybackslash}p{8.8cm}@{}}
        \toprule
        {\bf Relation} & {\bf Cited Claim} & {\bf Citing Claim} & {\bf Description}\\
        \midrule

        \cfsupport & Incorporating a small amount of human-labeled data into distant supervision significantly improves relation extraction performance.~\citep{pershina-etal-2014-infusion} & Crowdsourced annotation of training data can significantly improve performance for relation extraction compared to methods based solely on distant supervision.~\citep{liu-etal-2016-effective} & The citing claim provides evidence or results that are consistent with and reinforce the cited claim. \\
    
        \addlinespace
        
        \cfextend & GAN-BERT extends the fine-tuning of BERT-like architectures with unlabeled data in a generative adversarial setting.~\citep{croce-etal-2020-gan} & Extending transformer-based models with Semi-Supervised Generative Adversarial Networks (SS-GAN) improves performance on the Dialect Arabic Identification task.~\citep{yusuf-etal-2022-arabic} & The citing claim applies the cited claim to a new setting, task, dataset, or method while preserving its core assertion. \\
        \addlinespace
        
        \cfqualify & Neural network models, particularly those based on Long Short-Term Memory (LSTM), can achieve state-of-the-art performance on natural language inference tasks when trained on the SNLI corpus.~\citep{bowman-etal-2015-large} & The performance of high-performing NLI models is dramatically lower on a subset of examples considered 'hard' compared to the rest of the instances.~\citep{gururangan-etal-2018-annotation} & The citing claim restricts, conditions, or refines the scope of the cited claim without rejecting it. \\
        \addlinespace
        
        \cfrefute & Deeper models can extract more expressive features and improve performance in machine translation tasks.~\citep{bapna-etal-2018-training} & Increasing the depth of NMT models directly by stacking more blocks does not improve performance and can lead to optimization failure.~\citep{wu-etal-2019-depth} & The citing claim presents evidence or arguments that contradict the cited claim. \\
        \addlinespace
        
        \cfbackground & Byte pair encoding (BPE) is a suitable word segmentation strategy for neural network models.~\citep{sennrich-etal-2016-neural} & BPE-dropout is a simple and effective subword regularization method.~\citep{provilkov-etal-2020-bpe} & The cited claim is mentioned for context or positioning, without being evaluated or modified. \\

        \bottomrule
        
        \end{tabular}
    \end{adjustbox}
    }
\endgroup
    \caption{
    Illustrative examples of claim–claim relations in \dataset{}, showing how papers engage with prior claims.}
    \label{tab:claim_rel_exampl}
\end{table*}

\paragraph{Longitudinal Analyses of NLP Research.}
Several studies have analyzed the evolution of NLP and other scientific fields by examining trends in topics, methods, datasets, or contributions over time~\citep{hall-etal-2008-studying, xiao-etal-2022-datasets, pramanick-etal-2023-diachronic, pramanick-etal-2025-nature, abdalla-etal-2023-elephant, wahle-etal-2025-citation}. They provide valuable high-level perspectives and often rely on surface-level features such as keywords and metadata. Text mining and deep learning techniques have also been utilized to enable detailed analyses of the interactions among topics and their evolution~\citep{prabhakaran2016predicting, tan-etal-2017-friendships, salloum2017survey, hou-etal-2019-identification, pramanick-etal-2023-diachronic, sahinuc-etal-2024-efficient}. \dataset{} complements this line of work by introducing claims as the units of analysis, enabling finer-grained longitudinal analyses of how scientific ideas emerge, persist, and change.

\paragraph{Citation Analysis.}
A large body of work analyzes citations as signals of influence or scholarly communication, often using graphs or bibliometric measures~\citep{bornmann2008citation}. Moving beyond raw citation counts, prior work has proposed classifying citation intent or function, distinguishing roles such as background, use, or comparison~\citep{teufel-etal-2006-automatic, jurgens2018measuring, cohan2019structural}. These approaches operate primarily at the level of sentences and focus on rhetorical or functional aspects of citations. In contrast, our work models epistemic relations between scientific claims, which captures scientific engagement that cannot be recovered from citation intent labels alone.

\paragraph{Scientific Claim and Argument Mining.}
Prior work has explored scientific claim extraction, evidence identification, and argument structure modeling in research articles~\citep{al-khatib-etal-2021-argument, binder-etal-2022-full, flashner-etal-2025-scicompanion}. Closely related efforts in scientific fact verification and claim checking evaluate whether individual statements are supported by evidence, often within a single document or against external sources~\citep{wadden-etal-2020-fact, saakyan-etal-2021-covid}. While these approaches identify claims or argumentative components, they typically treat claims in isolation. Our work instead tracks interactions between claims across papers, explicitly modeling how later work engages with claims introduced in earlier publications. Table~\ref{tab:claim_rel_exampl} illustrates the claim-relation labels used in this work.



\section{\dataset{}: A Claim-Centric Dataset for Scientific Progress in NLP}
\label{sec:data}

To represent scientific progress at the level of individual claims, \dataset{} models how claims are expressed and engaged with across papers.

\subsection{Conceptual Framework}

\subsubsection{What is a Scientific Claim?}
In this work, we define a \textit{scientific claim} as a \textit{central, empirically testable assertion about methods, data, or phenomena that a paper advances as part of its contribution}. This definition aligns with established views of scientific progress as the proposal and evaluation of testable assertions~\citep{popper1959logic, kuhn1970structure, lakatos2014falsification, swales2014genre}.
We distinguish abstract claims from their textual realizations.
A \textit{claim text} refers to a statement that expresses a claim; the same claim may be stated multiple times and in different linguistic forms across sections.
In \dataset{}, multiple realizations are normalized into a canonical claim representation, which serves as the unit of annotation and analysis.

\subsubsection{What are Claim Relations?}

\paragraph{Definition and Scope.}
A claim relation specifies how a claim from a citing paper engages with a claim from a cited paper. To interpret this interaction, \dataset{} uses the citation context -- namely, the sentence containing the citation marker together with its immediately preceding and following sentences -- as localized textual evidence linking the two papers. We treat claim identification as an upstream process and focus on modeling interactions between claims once identified.

\paragraph{Components of Claim--Claim Interaction.}
As illustrated in Figure~\ref{fig:fig1}, determining the relation between two claims requires reasoning over three sources of information: the cited claim, the citing claim, and the local citation context. The cited claim defines the target of engagement, the citing claim expresses the new scientific assertion, and the citation context provides rhetorical evidence for how the prior work is discussed. None of these components alone is sufficient to determine the relation.
Determining whether this interaction constitutes support, extension, or qualification requires considering all three components jointly.

\paragraph{Textual Grounding.}
We distinguish \textit{citing and cited claim texts} from the underlying claims they express. Claim relations are defined between claims, but grounded in their textual realizations within papers. A citing claim may occur anywhere in the citing paper and is not necessarily located within the local citation context. When multiple citation contexts refer to the same cited paper, \dataset{} associates the claim pair with the context that most directly expresses the interaction between the claims.

\paragraph{Relation Taxonomy.}
We define a claim--claim relation taxonomy grounded in prior work on argumentation and scientific discourse~\citep{toulmin2003uses, lauscher-etal-2018-argument}. \dataset{} includes five relations:
\textcolor{claimflowsupport}{\texttt{support}}, \textcolor{claimflowextend}{\texttt{extension}}, \textcolor{claimflowqualify}{\texttt{qualification}}, \textcolor{claimflowrefute}{\texttt{refutation}}, and \textcolor{claimflowbackground}{\texttt{background}}.
They capture whether a work reinforces, generalizes, narrows, contradicts, or references a prior claim without substantive evaluation. Table~\ref{tab:claim_rel_exampl} lists definitions and examples.

\subsection{Dataset Curation}

\paragraph{Data Preparation.}
To study how scientific claims evolve in NLP, we curate a corpus of papers from the ACL Anthology~\citep{bird2008acl}. Our corpus consists of $1{,}617$ papers published between 1979 and 2025 from venues that belong to the ``ACL Events'' category.
\footnote{Major NLP conferences and workshops, including: ACL, NAACL, EMNLP, EACL, AACL, and Findings.}
We randomly sample papers with at least one paper per year.

We retrieve bibliographic metadata from \texttt{anthology.bib}.\footnote{\url{https://aclanthology.org/anthology.bib.gz}} We parse paper text with Docling~\citep{livathinos2025docling} and retrieve citation contexts via the Semantic Scholar API~\citep{lo2020s2orc}.
We retain only citation links between papers, resulting in a directed citation graph with $2{,}140$ edges.
This restriction ensures that all cited claims are available for annotation.

        
        
        
        
\begin{table}[t]
    \centering
\begingroup
\definecolor{cfBackgroundBg}{HTML}{EBF3FD}
\definecolor{cfBackgroundFg}{HTML}{357EDD}
\definecolor{cfSupportBg}{HTML}{E8F7F1}
\definecolor{cfSupportFg}{HTML}{19A974}
\definecolor{cfExtendBg}{HTML}{FFF4CC}
\definecolor{cfExtendFg}{HTML}{FFB700}
\definecolor{cfQualifyBg}{HTML}{EFEAF6}
\definecolor{cfQualifyFg}{HTML}{5E2CA5}
\definecolor{cfRefuteBg}{HTML}{FDE6E7}
\definecolor{cfRefuteFg}{HTML}{E7040F}
\newcommand{\cfrel}[3]{\begingroup\setlength{\fboxsep}{1.3pt}\colorbox{#1}{\textcolor{#2}{\scriptsize\textsc{#3}}}\endgroup}
\newcommand{\cfbackground}{\cfrel{cfBackgroundBg}{cfBackgroundFg}{background}}
\newcommand{\cfsupport}{\cfrel{cfSupportBg}{cfSupportFg}{support}}
\newcommand{\cfextend}{\cfrel{cfExtendBg}{cfExtendFg}{extend}}
\newcommand{\cfqualify}{\cfrel{cfQualifyBg}{cfQualifyFg}{qualify}}
\newcommand{\cfrefute}{\cfrel{cfRefuteBg}{cfRefuteFg}{refute}}
\newcommand{\cfbar}[2]{\makebox[1.25cm][l]{\textcolor{#1}{\rule{#2}{1.1mm}}}}
    \scalebox{0.90}{\small
        \begin{tabular}{@{}l r@{\hspace{0.55em}}l@{}}
        \toprule
        {\bf Relation} & \multicolumn{1}{c@{}}{{\bf Share ($\%$)}} \\
        \midrule
        \cfsupport & 20.9 & \cfbar{cfSupportFg}{0.50cm} \\
        \cfextend & 13.4 & \cfbar{cfExtendFg}{0.32cm} \\
        \cfqualify & 5.4 & \cfbar{cfQualifyFg}{0.13cm} \\
        \cfrefute & 2.1 & \cfbar{cfRefuteFg}{0.06cm} \\
        \cfbackground & 58.2 & \cfbar{cfBackgroundFg}{1.25cm} \\
        
        \bottomrule
        
        \end{tabular}
    }
\endgroup
    \caption{Percentages of relation types in \dataset{}.}
    \label{tab:rel_dist}
\end{table}

\paragraph{Annotation.}
Annotation was conducted by two NLP researchers in a two-stage workflow~\citep{cohan2019structural,teufel-etal-2006-automatic}.
\textbf{Claim identification.} First, the annotators are provided with the abstract, introduction, and conclusion of each paper and they identify \textit{claim texts} (statements expressing testable assertions about methods, data, or phenomena advanced by the paper). When necessary, annotators make minimal edits (e.g., resolving anaphora) to make the claim texts self-contained. For each claim text, we record its originating sections. \textbf{Relation labeling.} In the second phase, annotators examine papers that cite those annotated in the first phase. Annotators inspect the citation context in the citing paper together with the claim texts in both the citing and cited papers, and assign a claim relation label (from Table~\ref{tab:claim_rel_exampl}). Ambiguous cases are flagged and resolved through adjudication.

Both annotators label a shared subset of $1{,}001$ papers for agreement evaluation before independently annotating an additional $308$ papers each. The resulting corpus contains $1{,}617$ annotated papers (refer to Appendix~\ref{app:annot_instrn} for details). We refer to the resulting corpus of $1{,}617$ annotated papers as \dataset{}.

\paragraph{Agreement and Quality.}
We measure inter-annotator agreement using Krippendorff's $\alpha$ for claim identification, treated as a sentence-level binary labeling task with class imbalance, and Cohen's $\kappa$ for claim-relation annotation, where annotators assign a single label from a fixed set~\citep{krippendorff2018content, artstein2008inter}.
Table~\ref{tab:rel_iaa} shows $\alpha=0.75$ for claim identification and $\kappa=0.77$ for claim relation classification, comparable to prior work in scholarly document understanding~\citep{yang2018scidtb, hou2021tdmsci, lauscher2022multicite}. The corresponding $95\%$ confidence intervals range from $0.73$--$0.81$ for $\alpha$ and $0.72$--$0.82$ for $\kappa$ ($p<0.05$).

\subsection{Dataset Statistics}

\dataset{} consists of $1{,}617$ NLP research papers published between 1979 and 2025, annotated with $5{,}689$ scientific claims and $4{,}871$ claim relations inferred from citation contexts. Each paper contributes between 1 and 4 claims, with an average of $3.18$ claims per paper.

Claim relations are unevenly distributed across relation types. As shown in Table~\ref{tab:rel_dist}, \texttt{support} and \texttt{background} relations are most common, while \texttt{extend}, \texttt{qualify}, and \texttt{refute} are comparatively rare. The resulting claim graph is substantially sparser than the underlying citation graph, highlighting that citations do not necessarily correspond to substantive claim-level engagement~\citep{nicholson2021scite}.

    

\begin{table}[t]
    \centering
\begingroup
\definecolor{cfBackgroundBg}{HTML}{EBF3FD}
\definecolor{cfBackgroundFg}{HTML}{357EDD}
\definecolor{cfSupportBg}{HTML}{E8F7F1}
\definecolor{cfSupportFg}{HTML}{19A974}
\definecolor{cfExtendBg}{HTML}{FFF4CC}
\definecolor{cfExtendFg}{HTML}{FFB700}
\definecolor{cfQualifyBg}{HTML}{EFEAF6}
\definecolor{cfQualifyFg}{HTML}{5E2CA5}
\definecolor{cfRefuteBg}{HTML}{FDE6E7}
\definecolor{cfRefuteFg}{HTML}{E7040F}
\newcommand{\cfrel}[3]{\begingroup\setlength{\fboxsep}{1.3pt}\colorbox{#1}{\textcolor{#2}{\scriptsize\textsc{#3}}}\endgroup}
\newcommand{\cfbackground}{\cfrel{cfBackgroundBg}{cfBackgroundFg}{background}}
\newcommand{\cfsupport}{\cfrel{cfSupportBg}{cfSupportFg}{support}}
\newcommand{\cfextend}{\cfrel{cfExtendBg}{cfExtendFg}{extend}}
\newcommand{\cfqualify}{\cfrel{cfQualifyBg}{cfQualifyFg}{qualify}}
\newcommand{\cfrefute}{\cfrel{cfRefuteBg}{cfRefuteFg}{refute}}
    \scalebox{0.90}{\small
        \begin{tabular}{@{}l c@{}}
        \toprule
        {\bf Relation} & {\bf Cohen's $\kappa$} \\
        \midrule
        \cfsupport & 0.80 \\
        \cfextend & 0.73 \\
        \cfqualify & 0.72 \\
        \cfrefute & 0.78 \\
        \cfbackground & 0.82 \\
        \midrule
        {\bf Overall agreement} & {\bf 0.77} \\
        \bottomrule
    
        \end{tabular}}
\endgroup
    \caption{Relation-wise inter-annotator agreement.}
    \label{tab:rel_iaa}
\end{table}

    

\section{Automatic Claim Relation Classification}
\label{sec:task}

We define \textit{Claim Relation Classification} as the task of predicting how a citing claim engages with a cited claim and benchmark multiple models on \dataset{}.

\subsection{Task Definition}

We formulate the task as a supervised multi-class classification problem. Given a cited claim $h_{\text{cited}}$ from paper $\mathcal{B}$, a citing claim $h_{\text{citing}}$ from paper $\mathcal{A}$, and a citation context $s$ (defined as the sentence containing the citation marker and its immediately preceding and following sentences), the model predicts the epistemic relation
$r \in$ \{\textcolor{claimflowbackground}{\texttt{background}}, \textcolor{claimflowsupport}{\texttt{support}}, \textcolor{claimflowextend}{\texttt{extension}}, \textcolor{claimflowqualify}{\texttt{qualification}}, \textcolor{claimflowrefute}{\texttt{refutation}}\} given an input $(h_{\text{cited}}, s, h_{\text{citing}})$.

\subsection{Models and Baselines}

We evaluate a range of models to assess their ability to predict claim-relations. Our goal is not to exhaustively optimize architectures, but to compare representative modeling paradigms and identify which aspects of the task current models handle well or struggle with.

\paragraph{Encoder-Based Models.}
We evaluate transformer-based encoder models BERT~\citep{devlin-etal-2019-bert} and RoBERTa~\citep{liu2019roberta} as supervised baselines for claim relation classification.
We additionally evaluate DeBERTa-v3~\citep{he2021deberta}, a stronger encoder architecture with disentangled attention; and to assess the effect of domain-specific pretraining, we evaluate SciBERT~\citep{beltagy2019scibert}.
For each instance, we linearize the input triple $(h_{cited}, s, h_{citing})$ and feed the resulting sequence into a single encoder. The final hidden representation of the $[CLS]$ token is passed to a softmax classifier over the relation label set.

\paragraph{Large Language Models.}
We also evaluate large language models (LLMs) in a prompt-based setting.
For proprietary models, we evaluate GPT-3.5-turbo~\citep{achiam2023gpt}, GPT-4o, and GPT-4.1. For open-source models, we evaluate LLaMA-3-70B-Instruct~\citep{dubey2024llama} and Mixtral-8x7B-Instruct~\citep{jiang2024mixtral}.

\subsection{Experimental Setup}

\paragraph{Data Splits.}
We evaluate all models on \dataset{} using stratified train-validation-test splits constructed at the paper level to avoid leakage across citations.
Papers are partitioned into 70-15-15 splits, and label distributions are preserved across splits.

\paragraph{Training Details.}
We fine-tune encoder-based models using hyperparameters selected on the validation set (details in Appendix~\ref{app:exp_details}). LLMs are evaluated in zero-shot and four-shot prompting settings using a unified prompt template (details in Appendix~\ref{app:llm_prompts}).

\paragraph{Evaluation.}
Following \citet{uma2021learning}, for multi-class classification, we use macro-averaged precision, recall, and F1-score for label-based evaluation, aggregating performance across labels.

\subsection{Results and Error Analysis}

        
\begin{table}[t]
    \centering
\begingroup
\definecolor{modelBest}{HTML}{357EDD}
\definecolor{modelGroupBg}{HTML}{F2F2F2}
\newcommand{\modelsetting}[1]{\begingroup\setlength{\fboxsep}{1.2pt}\colorbox{modelGroupBg}{\scriptsize\textsc{#1}}\endgroup}
\newcommand{\bestscore}[1]{\textcolor{modelBest}{\bf #1}}
    \scalebox{0.80}{
\small
        \begin{tabular}{@{}l l c c c@{}}
        \toprule
        {\bf Setting} & {\bf Model} & {\bf P} & {\bf R} & {\bf F1}\\
        \midrule
        \modelsetting{Baseline} & Majority & 0.18 & 0.20 & 0.18 \\
        \midrule
        \multirow{4}{*}{\modelsetting{Fine-tuned}} & BERT & 0.51 & 0.50 & 0.50 \\
        & RoBERTa & 0.73 & 0.71 & 0.70 \\
        & SciBERT & 0.76 & 0.75 & 0.75 \\
        & DeBERTa & 0.77 & 0.76 & 0.76 \\ 
        \midrule
        \multirow{5}{*}{\modelsetting{Prompted}} & GPT-3.5-turbo & 0.75 & 0.75 & 0.75 \\
        & GPT-4o & 0.78 & 0.78 & 0.78 \\
        & GPT-4.1 & \textbf{0.82} & \textbf{0.81} & \textbf{0.81} \\
        & LLaMA-3-70B & 0.71 & 0.70 & 0.70 \\
        & Mixtral-8x7B & 0.73 & 0.74 & 0.73 \\
        \bottomrule
        
        \end{tabular}
    }
\endgroup
    \caption{Performance of different models for claim-relation classification.}
    \label{tab:eval_results_main}
\end{table}

\paragraph{Main Results.}
Table~\ref{tab:eval_results_main} shows the results.
Among fine-tuned encoders, RoBERTa consistently outperforms BERT, confirming the benefit of stronger pretrained representations. SciBERT further improves performance, suggesting that domain-specific pretraining on scientific text benefits modeling claim interactions. DeBERTa achieves the strongest encoder performance, indicating that improved contextual representation learning helps capture fine-grained claim relations. In contrast, the majority-class baseline performs substantially worse, showing that the task cannot be solved through label frequency alone.

In the zero-shot setting, all LLMs perform below most of the fine-tuned encoders, whereas few-shot prompting leads to consistent improvements across models. GPT-4.1 achieves the strongest overall performance among LLMs.
Using the full input $(h_{cited}, s, h_{citing})$ consistently improves performance over variants that omit the citation context or one of the claims (ablation studies in Appendix~\ref{app:add_results}).

\paragraph{Error Analysis.}
Qualitative inspection of model errors in Table~\ref{tab:error_qual} (in Appendix~\ref{app:add_results}) reveals two dominant failure modes. First, models frequently confuse extension with qualification, particularly when a citing paper applies a prior claim to a new setting while simultaneously introducing constraints. Second, models struggle with cases where support or refutation is expressed indirectly. Hedging, cautious language, and implicit comparisons can obscure epistemic stance, leading models to default to the background label. Overall, the results indicate that while current models can capture coarse scientific stance, they struggle with nuanced claim-level reasoning.

\subsection{\datasetauto{}}

To enable large-scale analysis beyond expert annotation, we construct an automatically inferred claim graph over $13{,}358$ papers from major ACL Anthology venues spanning 1979--2025 using a two-stage pipeline that combines automatic claim identification and claim relation classification (Figure~\ref{fig:autograph_pipeline} in Appendix~\ref{app:autograph_validation}).

\paragraph{Claim identification.}
First, we perform automatic claim identification at the sentence level to extract candidate claim texts from unannotated papers. Following \citet{pramanick-etal-2025-nature}, we fine-tune DeBERTa on \dataset{} and apply the resulting model to the larger ACL Anthology corpus, treating each detected sentence as an individual claim text.

\paragraph{Relation prediction.}
Next, for each citation between papers, we consider pairs of citing and cited claim texts and use GPT-4.1 in a few-shot setting to determine whether the citing claim engages with the cited claim and, if so, to predict the corresponding epistemic relation. When a paper cites another paper multiple times, each citation mention yields a separate citation context, defined as the citation sentence together with its immediately preceding and following sentences, and claim interactions are predicted independently for each context (details in Appendix~\ref{app:autograph_relation_prompt}).

\paragraph{Canonicalization.}
This process produces an automatically inferred directed claim graph in which nodes correspond to claim texts and edges represent predicted claim-claim relations. To reduce redundancy arising from repeated or near-duplicate claim texts, we apply a lightweight canonicalization procedure that clusters highly similar claim surface realizations and retains a single representative per cluster (Appendix~\ref{app:canonicalization}).

We note that the resulting claim graph serves as a scalable approximation that enables aggregate analyses of claim propagation and interaction patterns across NLP literature~\citep{teodorescu-mohammad-2023-evaluating}.

\subsubsection{Validation of \datasetauto{}}

We evaluate the automatic pipeline by comparing inferred claim graphs against expert annotations under two settings (details in Appendix~\ref{app:autograph_validation}): (i) relation classification with gold claim pairs and citation contexts, and (ii) full end-to-end graph recovery. For end-to-end evaluation, a predicted edge is correct only if the citing claim, cited claim, and relation label match the gold annotation under the canonicalization procedure.
Table~\ref{tab:end_to_end} reports end-to-end edge recovery performance.

\begin{table}[t]
    \centering
\begingroup
\newcommand{\validsetting}[1]{\begingroup\setlength{\fboxsep}{1.2pt}\colorbox{black!6}{\scriptsize #1}\endgroup}
\newcommand{\validscore}[1]{\textcolor{claimflowbackground}{\bf #1}}
    \scalebox{0.95}{\small
    \begin{tabular}{@{}lccc@{}}
        \toprule
            \textbf{Setting} & \textbf{P} & \textbf{R} & \textbf{F1} \\
            \midrule
            \validsetting{Gold claims + context} & 0.82 & 0.81 & 0.81 \\
            \validsetting{Automatic pipeline} & 0.81 & 0.79 & 0.80 \\
        \bottomrule
    \end{tabular}
    }
\endgroup
    \caption{Pipeline validation results for gold-claim classification and automatic edge recovery.}
    \label{tab:end_to_end}
\end{table}

Despite error propagation across multiple stages, the automatic pipeline achieves strong end-to-end performance, suggesting that meaningful claim interaction structure can be recovered automatically at scale. Aggregate analyses in \S~\ref{sec:analysis} also remain stable under realistic levels of pipeline noise.

\section{Analyzing Claim Flow in NLP}
\label{sec:analysis}

Using \datasetauto{}, we analyze the evolution of NLP research through three dimensions:
claim engagement by later work (\S~\ref{subsec:claim_engagement}), claim propagation across the literature (\S~\ref{subsec:claim_propagation}), and temporal claim evolution (\S~\ref{subsec:claim_evolution}).

\subsection{Claim Engagement}
\label{subsec:claim_engagement}

\noindent\textcolor[HTML]{00449E}{\textbf{Takeaway.}} \textit{NLP papers usually reuse prior claims as background or support. Challenges are rare, concentrated early in a claim's life cycle; these patterns are consistent across major venues.}

\begin{enumerate}[wide, noitemsep, labelindent=0pt, series="rqs", start=1, label={\textcolor{black!70}{\bf Q\arabic*.}}]
\item\textbf{How do NLP papers engage with claims from prior work?}
\end{enumerate}

\begin{figure}
    \centering
    \includegraphics[width=\columnwidth]{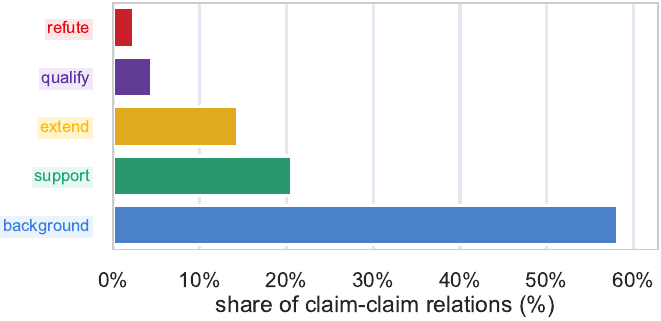}
    \caption{Distribution of claim–claim relations across \datasetauto{}.}
    \label{fig:rq1_claim_dist}
\end{figure}

\noindent We analyze the overall distribution of relation types across \datasetauto{}, which provides a coarse-grained view of how claims are typically used when cited by later work.

\paragraph{Results.} Figure~\ref{fig:rq1_claim_dist} shows that the majority of claim-claim interactions fall into the \textcolor{claimflowbackground}{\texttt{background}} ($58.2\%$) and \textcolor{claimflowsupport}{\texttt{support}} ($20.6\%$) categories, indicating that claims are most often reused as contextual premises or explicitly supported by subsequent work. \textcolor{claimflowextend}{\texttt{extend}} ($14.4\%$) occur less frequently, while \textcolor{claimflowqualify}{\texttt{qualify}} ($4.5\%$), and \textcolor{claimflowrefute}{\texttt{refute}} ($2.4\%$) are comparatively rare.

\paragraph{Discussion.} Results are consistent with long-standing norms in the NLP literature, where progress has typically been framed as cumulative rather than adversarial. Early symbolic and rule-based approaches~\citep{church-1988-stochastic} were rarely rejected outright; instead, they were later positioned as foundational context when probabilistic and corpus-driven methods gained prominence~\citep{brown-etal-1993-mathematics}. 
A similar dynamic accompanied the shift from feature-engineered statistical models to neural architectures, where prior methods were generally contextualized or subsumed rather than explicitly refuted~\citep{collobert2011natural}. 
Even during major paradigm transitions, such as the move from phrase-based statistical machine translation to neural machine translation, earlier approaches were more often bounded or implicitly superseded than directly contradicted~\citep{bahdanau2014neural}. 

\begin{enumerate}[wide, noitemsep, labelindent=0pt, resume="rqs", label={\textcolor{black!70}{\bf Q\arabic*.}}]
\item\textbf{How often are claims challenged in NLP research, and how are challenged claims subsequently engaged?}
\end{enumerate}

\begin{figure}[ht]
    \centering
    \includegraphics[width=\columnwidth]{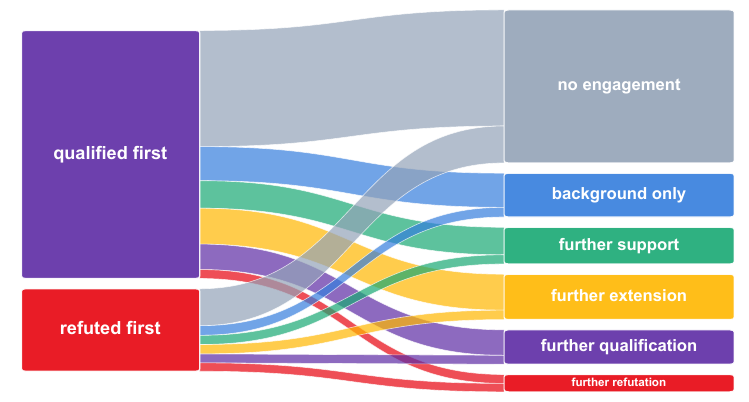}
    \caption{Distribution of post-challenge engagement for claims; flow widths indicate the number of claims.}
    \label{fig:rq3_challenge_sankey}
\end{figure}

\begin{figure}[ht]
    \centering
    \includegraphics[width=0.9\columnwidth]{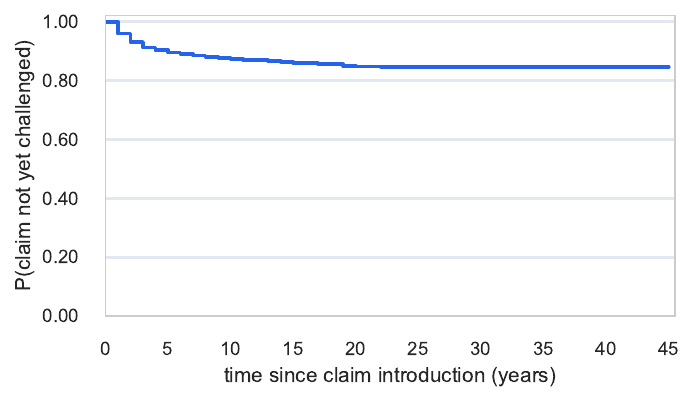}
    \caption{Survival curve for time to first claim challenge; claims not challenged during the observation window are right-censored.}
    \label{fig:rq3_challenge_survival_plot}
\end{figure}

\noindent We define \textit{challenge} as interactions labeled \textcolor{claimflowqualify}{\texttt{qualify}} or \textcolor{claimflowrefute}{\texttt{refute}}, and analyze when claims are first challenged and how 
subsequently engaged. 
For each claim $h$, 
we measure the time to first challenge, defined as the difference between the publication year of the claim's originating paper and the earliest subsequent paper that challenges it. To examine what happens after 
the challenge,
we identify all post-challenge interactions, i.e., claim--claim interactions occurring after the first 
challenge.

\paragraph{Results.} We find that only a small number of claims are ever challenged ($11.1\%$). 
Among challenged claims, \textcolor{claimflowqualify}{qualification} ($8.3\%$) is more common than direct \textcolor{claimflowrefute}{refutation} ($2.8\%$), indicating that claims are more often constrained or refined than rejected. Figure~\ref{fig:rq3_challenge_sankey} shows that after an initial challenge, many claims receive no further substantive engagement ($46.3\%$) or are cited only as \textcolor{claimflowbackground}{background} ($13.2\%$). Among claims that continue to be engaged, further \textcolor{claimflowsupport}{support} ($11.1\%$) and \textcolor{claimflowextend}{extension} ($13.8\%$) are more common than repeated \textcolor{claimflowqualify}{qualification} ($10.4\%$) or \textcolor{claimflowrefute}{refutation} ($5.2\%$), suggesting that challenges rarely lead to prolonged adversarial discourse. Figure~\ref{fig:rq3_challenge_survival_plot} shows that challenges usually occur early in a claim's life cycle; among claims that are challenged, the median time to first challenge is $2$ years.

\paragraph{Discussion.} Results show that NLP research engages with challenged claims in a limited and non-escalatory manner. The survival curve reflects the tendency toward early vetting followed by normalization. Once a claim is qualified or refuted, later work tends either to move on or to re-incorporate the claim in a modified form, rather than continuing to contest it. For example, early debates around parsing models led not to repeated rejection of specific claims~\citep{magerman-1995-statistical, collins-1997-three}, but to shifts in evaluation regimes and modeling constraints as statistical parsing matured~\citep{charniak-2000-maximum}. 


\begin{enumerate}[wide, noitemsep, labelindent=0pt, resume="rqs", label={\textcolor{black!70}{\bf Q\arabic*.}}]
\item \textbf{Do scientific claims in NLP research accumulate uncertainty over time?}
\end{enumerate}

\begin{figure}
    \centering
    \includegraphics[width=\columnwidth]{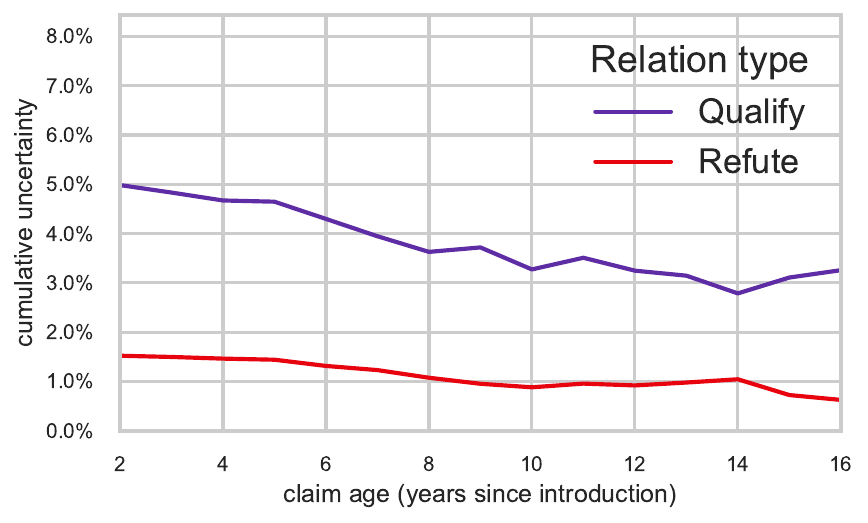}
    \caption{Cumulative uncertainty remains highest early in a claim's life cycle.}
    \label{fig:rq7_claim_uncertainty}
\end{figure}

\noindent To examine whether claims accumulate uncertainty over time, we analyze the temporal trajectories of claim-claim interactions for each cited claim. We operationalize uncertainty using relations: \textcolor{claimflowqualify}{qualify} and \textcolor{claimflowrefute}{refute}, which respectively indicate that a claim's scope is constrained or that it is contradicted by later work. 

For a claim $h$, we consider its sequence of downstream interactions ordered by publication year. We define the cumulative uncertainty at time $t$ as the proportion of all interactions up to $t$ that are labeled \textcolor{claimflowqualify}{qualify} and \textcolor{claimflowrefute}{refute}. We track how this proportion evolves as a function of claim age.

\paragraph{Results.} Figure~\ref{fig:rq7_claim_uncertainty} shows that cumulative uncertainty remains low throughout the claim lifecycle and is highest early in a claim's lifecycle and decreases over time. In the first few years after introduction, a non-trivial fraction of interactions involve \textcolor{claimflowqualify}{qualify} or \textcolor{claimflowrefute}{refute}, but as claims age, the proportion of uncertainty-inducing interactions declines, stabilizing at a low level for long-lived claims. Across all ages, uncertainty is dominated by \textcolor{claimflowqualify}{qualify}, while \textcolor{claimflowrefute}{refute} remains rare throughout.

\paragraph{Discussion.} The observed decline is consistent with uncertainty in NLP claims being front-loaded rather than cumulative. New claims may be tested, bounded, or occasionally contradicted as the community probes their validity and scope; claims that remain active at later ages consequently exhibit a lower proportion of uncertainty-inducing interactions.

Historically, this pattern aligns with how influential ideas in NLP have matured. Early claims -- such as those associated with new modeling paradigms or resources -- often face scrutiny shortly after introduction, during which limitations are identified and scope conditions clarified. Once these boundaries are established, subsequent work typically builds on the refined claim rather than re-challenging it. 

Taken together, these results indicate that uncertainty in NLP operates as an early-phase corrective mechanism rather than an ongoing process of erosion. Claims either stabilize after initial qualification or fade from active use, leaving long-lived claims with comparatively low cumulative uncertainty. This dynamic helps explain why explicit refutation remains rare and short-lived, and reinforces a broader picture of NLP research as progressing through early selection and path-dependent consolidation, rather than sustained adversarial debate.

\begin{enumerate}[wide, noitemsep, labelindent=0pt, resume="rqs", label={\textcolor{black!70}{\bf Q\arabic*.}}]
\item\textbf{Do publication venues differ in how scientific claims are engaged in NLP research?}
\end{enumerate}


\begin{figure}
    \centering
    \includegraphics[width=\columnwidth]{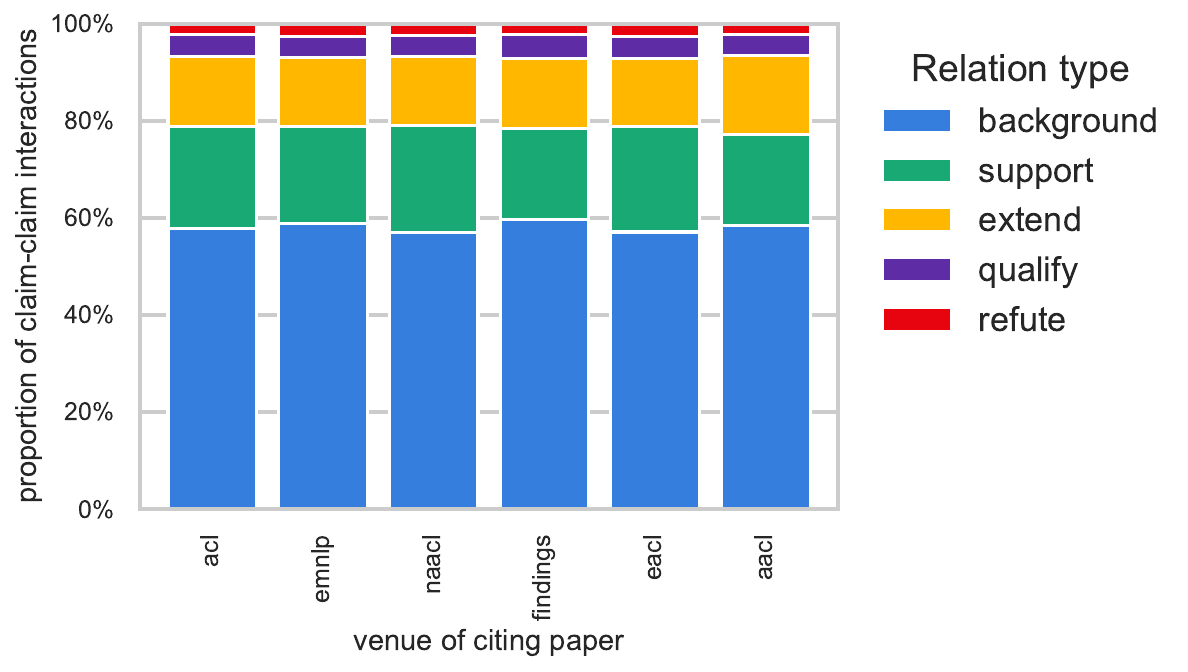}
    \caption{Distribution of claim--claim relation types across major NLP venues.}
    \label{fig:rq6_claim_rel_venue}
\end{figure}

\noindent To examine venue-level differences in claim engagement, we group claim-claim interactions by the publication venue of the \textit{citing paper}. For each venue $v$, we compute the distribution of relation types (\textcolor{claimflowsupport}{support}, \textcolor{claimflowextend}{extend}, \textcolor{claimflowqualify}{qualify}, \textcolor{claimflowrefute}{refute}, \textcolor{claimflowbackground}{background}) across all interactions originating from that venue. To ensure comparability across venues with different publication volumes, we report normalized proportions (normalized by the number of publications) rather than raw counts.

\paragraph{Results.} Figure~\ref{fig:rq6_claim_rel_venue} shows nearly consistent engagement patterns across ACL, EMNLP, NAACL, EACL, AACL, and Findings. In all venues, \textcolor{claimflowbackground}{background} relations dominate ($\approx57$--$60\%$), followed by \textcolor{claimflowsupport}{support} ($\approx19$--$22\%$) and \textcolor{claimflowextend}{extension} ($\approx14$--$16\%$). \textcolor{claimflowqualify}{Qualification} remains uncommon ($\approx4$--$5\%$), and \textcolor{claimflowrefute}{refutation} is rare ($\approx2$--$3\%$). The small differences at the margins do not alter the overall ordering of relation types.

\paragraph{Discussion.} The cross-venue similarity suggests that the aggregate engagement patterns are not driven by a single publication venue. Within the ACL Anthology venues studied here, venue choice modulates how claims are reused only at the margins and does not alter the dominant, cumulative mode of engagement.


\subsection{Claim Propagation}
\label{subsec:claim_propagation}

\noindent\textcolor[HTML]{00449E}{\textbf{Takeaway.}} \textit{Most claims remain locally used, but the few that propagate tend to be taken up quickly.}

\begin{enumerate}[wide, noitemsep, labelindent=0pt, resume="rqs", label={\textcolor{black!70}{\bf Q\arabic*.}}]
\item\textbf{Do claims propagate across NLP papers or remain isolated?}
\end{enumerate}
\begin{figure}[ht]
    \centering
    \includegraphics[width=\columnwidth]{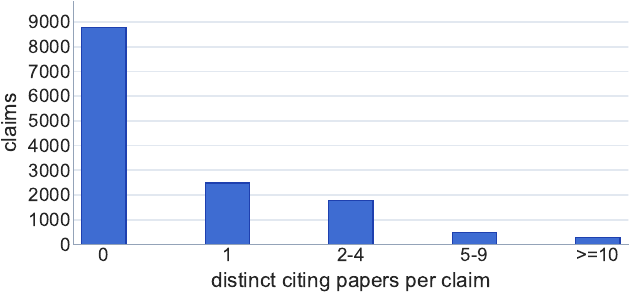}
    \caption{Percentage of papers reusing claims.}
    \label{fig:rq2_prop_dist}
\end{figure}

\begin{figure}[ht]
    \centering
    \includegraphics[width=0.9\columnwidth]{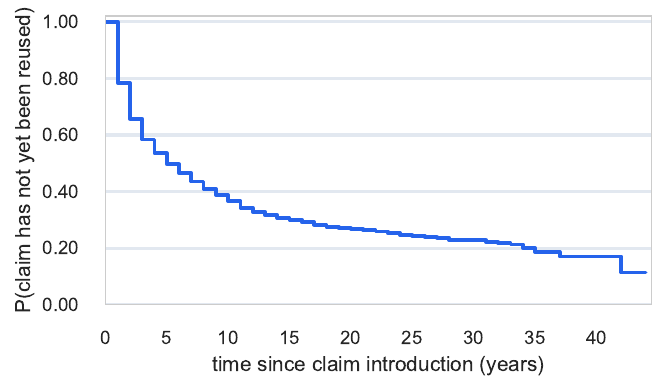}
    \caption{Survival curve for time to first claim reuse.}
    \label{fig:rq2_survival_plot}
\end{figure}

\noindent 
For each claim $h$, we compute its propagation count, defined as the number of unique papers that reference $h$. Claims with a propagation count of zero are considered \textit{isolated}, whereas claims referenced by one or multiple papers are considered \textit{propagated}. Additionally, to analyze temporal reuse, we measure the time between a claim's introduction and its first subsequent reuse, treating unreused claims as \textit{right-censored}.

\paragraph{Results.} In Figure~\ref{fig:rq2_prop_dist} we observe that most claims exhibit limited propagation across NLP papers, with $63.5\%$ claims remaining \textit{isolated}. Among claims that propagate, most are reused by only a small number of subsequent papers (mean propagation count $1.37$), while a minority propagate widely ($\sim$$2\%$ reused by ten or more papers).
Figure~\ref{fig:rq2_survival_plot} further shows that claim reuse is time-concentrated. The probability that a claim has not yet been reused drops sharply within the first few years after its introduction.

\paragraph{Discussion.} These engagement patterns show that influential ideas in NLP have often spread through rapid early uptake followed by consolidation, rather than gradual or sustained reuse. The early decline observed in the claim survival curve reflects this historical tendency toward early selection: claims that align with emerging datasets, tools, or modeling paradigms are quickly reused, whereas others effectively exit the active discourse without explicit refutation. 
For example, representational shifts -- such as the introduction of word embeddings and later contextualized representations -- were driven by a small number of claims that achieved early community adoption and subsequently dominated reuse patterns~\citep{mikolov2013efficient, peters-etal-2018-deep}, while alternative proposals introduced in the same period attracted little long-term engagement~\citep{collobert2008unified, turian-etal-2010-word}.



\begin{enumerate}[wide, noitemsep, labelindent=0pt, resume="rqs", label={\textcolor{black!70}{\bf Q\arabic*.}}]
\item\textbf{Do earlier claims shape later NLP research disproportionately?}
\end{enumerate}

\noindent For each claim $h$ we define its temporal position as \(\texttt{AgeRank(h)} = \frac{|\{h' \in \mathcal{H}: y(h')<y(h)\}|}{|\mathcal{H}|}\), where $y(h)$ is the publication year of the paper introducing $h$, and lower values correspond to earlier claims. Additionally, we measure \textit{claim influence} using 
the number of distinct subsequent papers that engage with a claim, normalized by its potential exposure time
\(\texttt{NormInfluence(h)}=\frac{\#\text{distinct papers citing } h}{\#\text{papers published after } y(h)}\). This normalization controls for the fact that earlier claims have more opportunities to be cited than later ones. We analyze the association between \texttt{AgeRank(h)} and \texttt{NormInfluence(h)} using rank correlations.

\paragraph{Results.} We observe a negative association between temporal position and normalized downstream influence. Specifically, the Spearman rank correlation between AgeRank and NormInfluence is $\rho=-0.495$, indicating that earlier claims receive higher normalized engagement from subsequent papers compared to later claims. This effect persists after normalizing for exposure time, suggesting that the observed advantage of earlier claims is not solely due to having more opportunities to be cited.

\paragraph{Discussion.} Results imply that claims introduced during periods of infrastructural or evaluative formation of NLP tend to exert long-lasting influence. For example, early claims associated with automatic evaluation metrics -- most notably BLEU -- rapidly shaped how machine translation progress was assessed and continued to structure downstream research long after their introduction~\citep{papineni-etal-2002-bleu}.

\subsection{Claim Evolution}
\label{subsec:claim_evolution}

\noindent\textcolor[HTML]{00449E}{\textbf{Takeaway.}} \textit{The claim graph stays sparse overall, yet increasingly organizes into thematic clusters that later partially reconnect through shared modeling paradigms.}

\begin{enumerate}[wide, noitemsep, labelindent=0pt, resume="rqs", label={\textcolor{black!70}{\bf Q\arabic*.}}]
\item\textbf{Does the claim graph densify or stratify over time in NLP research?}
\end{enumerate}

\begin{figure}[t]
    \centering

    \includegraphics[width=\columnwidth]{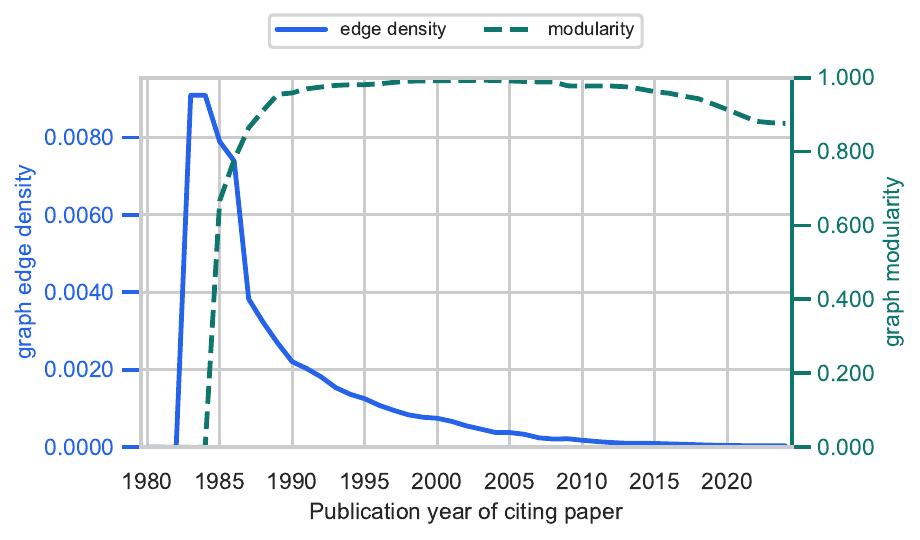}
    \caption{Structural evolution of the claim–claim interaction graph in NLP research.}
    \label{fig:rq5_structural_evolution}
\end{figure}

\noindent We construct a cumulative yearly claim--claim graph and track two structural properties: edge density, measuring how broadly claims connect, and modularity, measuring how strongly claims cluster into sub-communities. Formal definitions, separate density and modularity plots, and additional interpretation are provided in Appendix~\ref{subsec:structural_evolution_app}.


\paragraph{Results and Discussion.} Figure~\ref{fig:rq5_structural_evolution} shows that the edge density of the claim--claim graph remains extremely low overall (on the order of $10^{-3}$), after an initial rise in the early years attributable to small-graph effects. Despite continued growth in the number of claims and relations, density stabilizes and slightly declines over time, indicating that new claims engage with only a limited subset of prior claims. At the same time, modularity increases steadily before stabilizing and declining in recent years. This suggests that NLP claims are increasingly organized around specialized task and method clusters, but have partly reconnected under shared modeling paradigms such as pretrained and foundation models.

A complementary analysis of the claim-level structural roles underlying these aggregate graph trends is provided in Appendix~\ref{subsec:claim_dynamics}.

\subsection{Case Studies of Scientific Claim Evolution}
\label{subsec:case_study}

To complement the aggregate analysis, we examine two influential claim trajectories.
\textbf{BERT}~\citep{devlin-etal-2019-bert}. Following its introduction, the original claim -- deep bidirectional pretraining substantially improves downstream NLP performance -- rapidly accumulated support and extension interactions as later work adapted pretrained representations to new tasks and domains. Over time, interactions increasingly qualified the claim through concerns about reasoning ability, efficiency, and computational cost~\citep{kassner-schutze-2020-negated, ribeiro-etal-2020-beyond} (Figure~\ref{fig:case_study_trajectories}(a)). Rather than being overturned, the claim was progressively refined as the field evolved. \textbf{BLEU}~\citep{papineni-etal-2002-bleu}. Initially, the original claim -- that BLEU provides an effective automatic proxy for machine translation quality -- received strong support and widespread reuse as a standard evaluation metric in machine translation. Over time, later work increasingly qualified or refuted the claim by identifying limitations in semantic adequacy and correlation with human judgment~\citep{novikova-etal-2017-need, sellam-etal-2020-bleurt} (Figure~\ref{fig:case_study_trajectories}(b)). Despite sustained criticism, the metric remained deeply embedded in NLP evaluation practice. We provide additional qualitative studies in Appendix~\ref{app:case_studies}.


\section{Discussion}

\paragraph{Claim-level propagation.} Taken together, our findings suggest that scientific influence in NLP is unevenly distributed and only partially reflected by paper-level citation counts. Most claims remain isolated, whereas the relatively few claims that propagate tend to receive attention soon after publication. Earlier claims also receive greater normalized downstream engagement, even after accounting for their longer exposure to subsequent work. These results point to a process of selective consolidation in which a small set of claims becomes embedded in later research while many others leave little visible trace in the claim graph.

\paragraph{Relation dynamics and case studies.} The relation-level analysis further clarifies how this consolidation occurs. Explicit challenges are rare and usually arise early in a claim's life cycle; when influential claims remain active, later work more often extends or qualifies them than directly confirms or refutes them. The BERT and BLEU case studies illustrate two forms of this process: a modeling claim can retain its broader influence while its scope is progressively refined, and an evaluation claim can remain infrastructurally important despite accumulating qualifications and occasional refutations. Scientific change in NLP therefore appears less as a sequence of clean replacements than as continued reuse under revised scope, evidence, and methodological context.

\paragraph{Field-level structure.} At the field level, these local dynamics produce a claim graph that remains globally sparse while organizing into increasingly distinct clusters, followed by recent signs of reconnection around shared modeling paradigms. This pattern should not be interpreted as evidence that particular benchmarks, framing choices, or paradigms causally determine the direction of the field; our analyses do not directly test those mechanisms. Rather, it shows that claim-level interactions expose forms of uptake, refinement, and contestation that are obscured when citations or papers are treated as undifferentiated units. In this sense, \dataset{} complements citation-based scientometrics with a representation of scientific influence as interaction among specific assertions.

\section{Conclusion}

We introduced \dataset{}, the first expert-annotated dataset linking claims with labeled epistemic relations, together with the task of \textit{Claim Relation Classification} and an automatically inferred claim graph spanning decades of NLP research. Our analyses show that most claims remain isolated, explicit challenges are rare and short-lived, early claims exert disproportionate influence, and influential claims are more often reshaped through qualification and extension than simply confirmed or rejected. By making these interactions explicit, \dataset{} provides a foundation for studying scientific knowledge as an evolving system of interconnected claims.

\section{Applications and Future Work}

\paragraph{Applications.} The expert-annotated \dataset{} provides a structured benchmark for evaluating whether NLP models can identify how scientific claims relate to one another. Claim Relation Classification is complementary to citation-intent classification and natural language inference: it focuses specifically on the epistemic stance that a later scientific claim takes toward an earlier one, grounded in its citation context. The results in \S~\ref{sec:task} show that this form of reasoning is tractable, while persistent confusions between extension and qualification leave substantial room for better models.

The automatically inferred \datasetauto{} claim graph also creates opportunities for tools that operate over the scientific literature. Claim-centered navigation systems could help researchers follow how an assertion is reused, extended, bounded, or challenged across papers instead of returning only documents connected by citations. At an aggregate level, the graph could support further scientometric studies of idea diffusion, emerging research threads, and claims whose downstream engagement is unusually broad or contested. These are prospective applications rather than capabilities evaluated in this work, but the analyses presented here establish the claim-level structure needed to investigate them.

\paragraph{Future Work.} An immediate direction is to expand \dataset{} beyond ACL venues and NLP, allowing comparisons of claim evolution across publication communities and scientific domains. Broader coverage should include claims expressed throughout full papers, not only in abstracts, introductions, and conclusions, and should examine how corpus and venue boundaries affect the observed propagation patterns.

Improving automatic claim identification, relation prediction, and canonicalization would enable larger and more reliable claim graphs. Such graphs could be integrated into scholarly search and LLM-based research assistants, provided that predicted relations remain traceable to their source claims and citation contexts. Finally, combining claim-flow structure with evidence about model performance, robustness, datasets, and benchmarks could test whether early interaction patterns predict which claims persist or how research trajectories develop. We leave these predictive and causal questions to future work.

\section*{Limitations}

\paragraph{Corpus scope.}
This study focuses on NLP papers drawn from the ACL Anthology, specifically venues under the ACL Events umbrella. While these venues represent a central and influential subset of NLP research, important contributions also appear in AI Conferences, regional venues, and preprint platforms. As a result, the dataset does not capture the full breadth of scientific claims in NLP, and extending claim-level analysis beyond the ACL Anthology remains an important direction for future work.

\paragraph{Temporal coverage.}
Our longitudinal analyses aggregate over claims and relations within this curated corpus. Claims may continue to evolve beyond the temporal and venue boundaries of the dataset, and some influential ideas may be underrepresented due to sampling scope. Consequently, the trends we report should be interpreted as indicative patterns within a high-quality sample rather than an exhaustive account of the field. 

\paragraph{Annotation scope.}
In addition, our annotations focus on abstracts, introductions, and conclusions, where authors typically articulate their central claims most explicitly. While claims may also appear in the main body of papers, annotating full texts would require substantially greater effort. Incorporating claims from the full paper content remains a future extension of this work.

\paragraph{Automatic labeling.}
Finally, our large-scale analyses rely on models trained on the human-annotated dataset to automatically identify claim relations in a broader corpus. It is important to acknowledge that no model achieves perfect accuracy, which can impact the quality of such analyses. However, \citet{teodorescu-mohammad-2023-evaluating} showed that trend-level conclusions derived from large automatically labeled datasets tend to be highly accurate and show a strong correlation with trends identified through gold-label analysis, supporting the reliability and accuracy of our analyses. 

\section*{Ethics Statement}

\paragraph{Data and privacy.}
In this work, we use publicly available research papers from the ACL Anthology and do not involve personal data or human subjects. All annotations were performed on scholarly text and reflect interpretations of scientific claims rather than judgments about individual authors or research communities. 

\paragraph{Responsible interpretation.}
We acknowledge that identifying and relating scientific claims involves interpretive decisions, and that scientific progress cannot be reduced to quantitative signals alone. The analyses enabled by \dataset{} are intended to support reflective examination of research trends, not to prescribe scientific value or replace expert judgment. Any conclusions drawn from this work should therefore be considered in conjunction with qualitative, ethical, and social considerations that shape scientific practice. 

\bibliography{custom}

\appendix







\section{Additional Analysis}
\label{app:add_analysis}

\subsection{Claim-Level Structural Roles}
\label{subsec:claim_dynamics}

\begin{enumerate}[wide, noitemsep, labelindent=0pt, series="app-rqs", start=1, label={\textcolor{black!70}{\bf Q\arabic*.}}]
\item \textbf{Do scientific claims primarily attract downstream convergence or integrate multiple prior claims?}
\end{enumerate}

\noindent We analyze the structural roles of claims in the directed claim-claim interaction graph, where nodes correspond to claims and direct edges represent interactions from citing to cited claims. For each claim $h$, we compute its \textit{in-degree} $k^{in}(h)$, as the number of distinct downstream claims that reference $h$, and its \textit{out-degree} $k^{out}(h)$, as the number of distinct claims that $h$ references.

We characterize \textit{downstream convergence points} as claims with high in-degree and relatively low out-degree, indicating that multiple later claims converge on the same prior claim. Conversely, claims with high out-degree and relatively low in-degree \textit{integrate multiple antecedents}: they draw on several prior claims while attracting fewer downstream claim links~\citep{borgatti2000models, guimera2005functional}. To summarize this balance, we compute a signed degree-balance score for each claim \(D(h) = \frac{k^{out}(h) - k^{in}(h)}{k^{out}(h) + k^{in}(h) + \epsilon}\), where negative values indicate convergence-dominated claims and positive values indicate integration-dominated claims. We set $\epsilon=1$ to avoid division by zero and to assign neutral scores to isolated claims with no incoming or outgoing interactions. Because this smoothing also pulls low-degree claims toward zero, we use $D(h)$ as a descriptive balance measure rather than a calibrated centrality score.

\begin{figure}[b]
    \centering
    \includegraphics[width=\columnwidth]{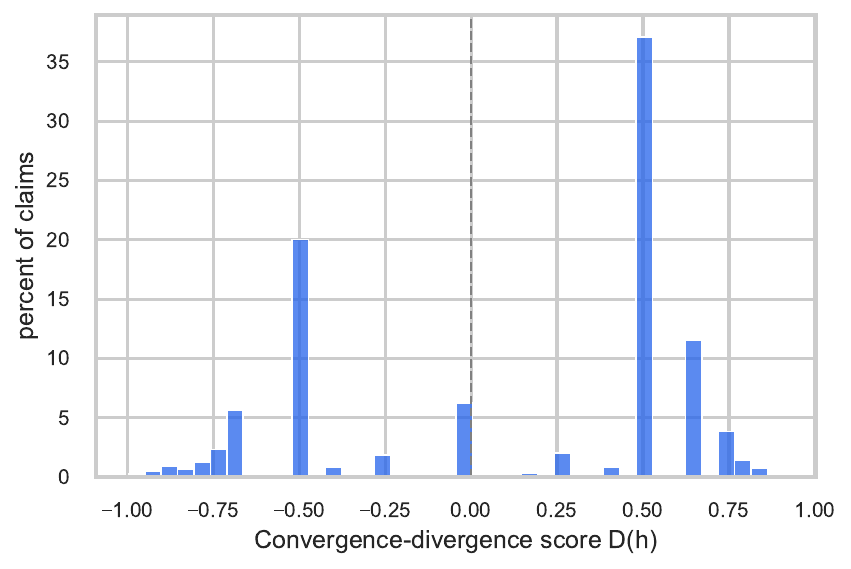}
    \caption[Distribution of signed degree-balance scores.]{Distribution of signed degree-balance scores. Positive values correspond to integration of multiple antecedents.}
    \label{fig:rq8_claim_convergence}
\end{figure}

\paragraph{Results.} Figure~\ref{fig:rq8_claim_convergence} shows the distribution of $D(h)$. While substantial mass lies near zero, indicating weakly connected or comparatively balanced claims, the distribution is skewed toward positive values. Thus, claims more often integrate multiple prior claims than attract many downstream claim links. The negative tail nevertheless shows that downstream convergence points also exist, though they are fewer in number.

\paragraph{Discussion.} This asymmetry suggests that NLP research exhibits a division of epistemic labor between integration and consolidation. Many claims synthesize and recombine existing ideas to explore new modeling choices, datasets, or problem formulations. Such integration is common during periods of rapid expansion, when researchers draw on multiple prior strands without immediately producing new focal claims -- for example, during the early statistical turn, when diverse probabilistic assumptions and feature combinations were explored in parallel~\citep{manning1999foundations}.

At the same time, the smaller but substantial mass of convergence-dominated claims reflects moments of field-wide consolidation, where a limited set of claims becomes shared reference points for subsequent work. Claims associated with major resources or evaluation standards -- such as the Penn Treebank~\citep{marcus-etal-1993-building} or widely adopted automatic evaluation metrics~\citep{papineni-etal-2002-bleu} -- illustrate this pattern, attracting sustained downstream engagement while themselves referencing relatively few predecessors. More recently, pretrained language models similarly functioned as convergence hubs, rapidly drawing engagement from across tasks and subfields~\citep{devlin-etal-2019-bert}.

Together, these findings indicate that NLP knowledge growth is neither uniformly integrative nor purely convergent. Instead, many claims combine several antecedents, punctuated by convergence onto a smaller number of stabilizing ideas.

\FloatBarrier

\section{Additional Experimental Details}
\label{app:exp_details}

\subsection{Training and Hyperparameter}

During fine-tuning of encoder-based models, we perform grid search over epochs
$e \in \{1,2,3,4,5\}$ and learning rates
$lr \in \{1\cdot10^{-4}, 5\cdot10^{-5}, 1\cdot10^{-5}\}$ using a batch size of $32$.
Hyperparameters are selected based on validation performance.

For LLMs, we evaluate both zero-shot and four-shot prompting settings using a unified prompt template. Few-shot examples are sampled from the training split and held constant across models. Following common practice in in-context learning evaluation~\citep{min-etal-2022-rethinking}, four-shot prompting is used as a lightweight demonstration setting rather than an exhaustive coverage of all relation types.

Each experiment is repeated three times. Across runs, performance variance remains below $0.02$ macro-F1 for all models.

\section{Annotation Instructions}
\label{app:annot_instrn}

This section provides detailed annotation guidelines used to construct \dataset{}. The guidelines expand the conceptual framework described in \S~\ref{sec:data} and specify the procedures used for claim identification, claim normalization, and claim-relation annotation.  


\subsection{Core Definitions}

Annotators were first presented with the following definitions, which guided all subsequent annotation decisions. 

\begin{tcolorbox}[
  colback=claimflowbackground!4,
  colframe=claimflowbackground!45!black,
  boxrule=0.4pt,
  arc=2mm,
  left=6pt,
  right=6pt,
  top=5pt,
  bottom=5pt
]

\paragraph{Scientific Claim.} A \textit{scientific claim} is a central, empirically testable assertion about methods, data, or phenomena that a research paper advances as part of its contribution.\par

\paragraph{Claim Text.} A \textit{claim text} is a span of text in a research paper that explicitly expresses a scientific claim. The same underlying claim may be expressed by multiple claim texts within a paper.

\end{tcolorbox}

Annotators were instructed to distinguish between \textit{claims} as abstract scientific commitments and \textit{claim texts} as their textual realizations. 

\subsection{Claim Identification}

Annotators were provided with the \textit{abstract}, \textit{introduction}, and \textit{conclusion} of each paper and instructed to identify claim texts from these sections. For each candidate sentence, annotators answered the following question:

\noindent\textit{Does this text express a scientific claim according to the definition above?}

\noindent A text was considered a valid claim text if it: 
    \begin{itemize}
        \item makes a testable assertion (rather than a definition or background statement),
        \item concerns a method, dataset, or empirical phenomenon,
        \item and reflects a central contribution emphasized by the authors.
    \end{itemize}

\noindent When necessary, annotators were allowed to make \textit{minimal edits} (e.g., resolving anaphora or removing references such as ``this model'') to ensure that claim texts were self-contained and interpretable in isolation. Annotators also recorded the section in which each claim text appeared. The citing claim text is not required to appear within the citation context itself and could occur elsewhere in the citing paper. 

\paragraph{Examples of Valid and Invalid Claims.}
Examples of valid claim texts include statements such as:
\begin{itemize}
    \item ``BERT improves performance on multiple NLP benchmarks.''
    \item ``Scaling model size improves few-shot reasoning performance.''
\end{itemize}

Examples that were not annotated as claims include:
\begin{itemize}
    \item purely descriptive background statements,
    \item definitions without empirical assertions,
    \item or methodological details not presented as central contributions. 
\end{itemize}

\subsection{Claim Normalization}
Multiple textual realizations within a paper may express the same underlying scientific claim. Annotators were instructed to group such realizations into a single canonical claim representation whenever the statements were semantically equivalent. Minor lexical, syntactic, or discourse-level variations were not treated as distinct claims.

\subsection{Claim-Relation Annotation}

In the second phase, annotators labeled relations between claims across papers. 

Annotators were shown 
\begin{itemize}
    \item a \textit{cited claim text} from a previously annotated paper,
    \item a \textit{citing claim text} from a later paper,
    \item and the \textit{citation context} in the citing paper (the sentence containing the citation marker together with the immediately preceding and following sentences)
\end{itemize}
When multiple citation contexts referred to the same cited paper, annotators selected the context that most directly expressed the interaction between the citing and cited claims.

Annotators then answered the following question:

\noindent\textit{Does the citing claim text substantively engage with the cited claim?}

\noindent Substantive engagement was defined as explicitly evaluating, building upon, qualifying, or challenging the content of the cited claim. 

When engagement was present, annotators assigned \textit{exactly one} of the following relation labels:

\begin{itemize}
    \item \textcolor{claimflowsupport}{\texttt{support}}: The citing claim provides evidence or results that reinforce the cited claim. 
    \item \textcolor{claimflowextend}{\texttt{extend}}: The citing claim generalizes or applies the cited claim to new settings, data, or tasks.
    \item \textcolor{claimflowqualify}{\texttt{qualify}}: The citing claim restricts the scope of the cited claim or specifies conditions under which it holds. 
    \item \textcolor{claimflowrefute}{\texttt{refute}}: The citing claim presents evidence or arguments that contradict the cited claim.
    \item \textcolor{claimflowbackground}{\texttt{background}}: The cited claim is referenced for context or comparison without substantive evaluation.
\end{itemize}

Annotators were instructed to base their decisions on the explicit content of the claim texts and citation context, not on external knowledge. 

\subsection{Adjudication and Edge Cases}

Ambiguous cases -- such as partial engagement, mixed signals, or unclear scope -- were flagged during annotation. Such cases were resolved through discussion, following the annotation guidelines and task definition described above.

Annotators were instructed to prioritize:

\begin{itemize}
    \item explicit textual evidence over implicit assumptions, 
    \item the claims under consideration rather than the overall contributions of a paper,
    \item and conservative labeling when evidence was insufficient to justify a stronger relation.
\end{itemize}

\subsection{Notes on Scope}

Annotators were informed that the goal of the dataset is to make claim-level scientific interaction observable. As a result, the annotation focuses on clear claims and clear instances of engagement, rather than marginal or purely rhetorical references.

\section{Prompting Large Language Models}
\label{app:llm_prompts}

\subsection{Task Framing}

Each instance consists of a cited claim text, a citing claim text, and the citation context from the citing paper. Models are instructed to select exactly one relation label from the predefined set:
\texttt{support}, \texttt{extend}, \texttt{qualify}, \texttt{refute}, or \texttt{background}. Models are explicitly instructed to output only the relation label, without explanations or additional text.

\subsection{Base Prompt Template}

All LLMs are prompted using the same base template to ensure consistency across models. The prompt is shown in Figure~\ref{fig:base_prompt}.







\begin{figure}[htbp]
\centering
\begin{tcolorbox}[
  colback=claimflowbackground!4,
  colframe=claimflowbackground!45!black,
  boxrule=0.5pt,
  arc=2mm,
  left=6pt,
  right=6pt,
  top=6pt,
  bottom=6pt,
  width=\linewidth
]
\small
\textbf{Instruction:} You are given two scientific claims from NLP research papers and a citation context from the citing paper.

\medskip
\textbf{Cited claim:}  
\textit{\{cited\_claim\}}

\medskip
\textbf{Citing claim:}  
\textit{\{citing\_claim\}}

\medskip
\textbf{Citation context:}  
\textit{\{citation\_context\}}

\medskip
\textbf{Task:}  
Determine how the citing claim engages with the cited claim.

\medskip
\textbf{Relation labels:}
\begin{itemize}[leftmargin=*, itemsep=2pt]
    \item \textcolor{claimflowsupport}{\texttt{support}}: the citing claim provides evidence that reinforces the cited claim.
    \item \textcolor{claimflowextend}{\texttt{extend}}: the citing claim generalizes or applies the cited claim to new settings or data.
    \item \textcolor{claimflowqualify}{\texttt{qualify}}: the citing claim restricts the scope or conditions under which the cited claim holds.
    \item \textcolor{claimflowrefute}{\texttt{refute}}: the citing claim contradicts the cited claim using opposing evidence or arguments.
    \item \textcolor{claimflowbackground}{\texttt{background}}: the cited claim is mentioned for context without evaluation.
\end{itemize}

\medskip
\textbf{Answer:}  
Choose exactly one label from the list above. Output only the label.
\end{tcolorbox}
\caption{Base prompt used for LLM-based claim relation classification.}
\label{fig:base_prompt}
\end{figure}

\subsection{Prompt for Automatic Graph Construction}
\label{app:autograph_relation_prompt}

For constructing \datasetauto{}, we use GPT-4.1 in a few-shot setting to classify each candidate citing--cited claim pair under a local citation context. The prompt follows the same label definitions as Figure~\ref{fig:base_prompt}, but makes the graph-construction decision explicit: if the citing claim does not significantly engage with the cited claim in the given context, the model must choose \texttt{irrelevant}. The template is shown in Figure~\ref{fig:autograph_relation_prompt}.

\begin{figure}[htbp]
\centering
\begin{tcolorbox}[
  colback=claimflowbackground!4,
  colframe=claimflowbackground!45!black,
  boxrule=0.5pt,
  arc=2mm,
  left=6pt,
  right=6pt,
  top=6pt,
  bottom=6pt,
  width=\linewidth
]
\small
\textbf{Instruction:} You are constructing a claim-level graph of NLP research. Given a cited claim, a citing claim, and the citation context, first decide whether the citing claim significantly engages with the cited claim.

\medskip
\textbf{Cited claim:}  
\textit{\{cited\_claim\}}

\medskip
\textbf{Citing claim:}  
\textit{\{citing\_claim\}}

\medskip
\textbf{Citation context:}  
\textit{\{citation\_context\}}

\medskip
\textbf{Decision rule:}  
Use the citation context as local evidence for the interaction. If the citing claim does not significantly engage with the cited claim in this context, output \texttt{irrelevant}; do not force a \texttt{background}, \texttt{support}, \texttt{extend}, \texttt{qualify}, or \texttt{refute} label. If significant engagement is present, choose the relation that best describes how the citing claim engages with the cited claim.

\medskip
\textbf{Relation labels:}
\begin{itemize}[leftmargin=*, itemsep=2pt]
    \item \textcolor{claimflowsupport}{\texttt{support}}: the citing claim provides evidence that reinforces the cited claim.
    \item \textcolor{claimflowextend}{\texttt{extend}}: the citing claim generalizes or applies the cited claim to new settings or data.
    \item \textcolor{claimflowqualify}{\texttt{qualify}}: the citing claim restricts the scope or conditions under which the cited claim holds.
    \item \textcolor{claimflowrefute}{\texttt{refute}}: the citing claim contradicts the cited claim using opposing evidence or arguments.
    \item \textcolor{claimflowbackground}{\texttt{background}}: the cited claim is mentioned for context without evaluation.
\end{itemize}

\medskip
\textbf{Few-shot examples:}  
\textit{\{four\_training\_examples\}}

\medskip
\textbf{Answer:} Output only the label.
\end{tcolorbox}
\caption{Prompt template used for relation prediction when constructing \datasetauto{}.}
\label{fig:autograph_relation_prompt}
\end{figure}

\subsection{Decoding and Post-processing}

Models are decoded using deterministic or low-temperature settings to reduce output variability. If a model produces output that does not exactly match one of the valid labels, the output is mapped to the closest valid label when unambiguous; otherwise, the instance is marked as incorrect. No external tools, retrieval mechanisms, or chain-of-thought prompting are used.

\section{Additional Validation of \datasetauto{}}
\label{app:autograph_validation}

We provide additional analyses assessing the reliability of \datasetauto{} in end-to-end inference settings, including evaluations of automatic claim identification, lightweight canonicalization, and robustness to graph perturbations. 

\begin{figure*}[t]
\centering
\resizebox{\textwidth}{!}{%
\begin{tikzpicture}[
    font=\small,
    stage/.style={
        draw=claimflowbackground!55!black,
        fill=claimflowbackground!6,
        rounded corners=2pt,
        line width=0.45pt,
        align=center,
        minimum height=1.25cm,
        minimum width=2.55cm,
        inner sep=4pt
    },
    note/.style={
        draw=black!20,
        fill=black!3,
        rounded corners=2pt,
        align=center,
        minimum height=0.62cm,
        minimum width=2.35cm,
        inner sep=2pt,
        font=\scriptsize
    },
    flowarrow/.style={
        -{Latex[length=2.2mm,width=1.5mm]},
        line width=0.55pt,
        draw=claimflowbackground!70!black
    },
    relationedge/.style={
        -{Latex[length=1.7mm,width=1.1mm]},
        line width=0.55pt
    }
]
\node[stage] (papers) {\textbf{ACL Anthology}\\papers\\{\scriptsize 1979--2025}};
\node[stage, right=1.0cm of papers, yshift=0.82cm] (claims) {\textbf{Claim}\\identification\\{\scriptsize sentence-level}};
\node[stage, right=1.0cm of papers, yshift=-0.82cm] (contexts) {\textbf{Citation}\\contexts\\{\scriptsize cited/citing papers}};
\node[stage, right=1.05cm of claims, yshift=-0.82cm] (pairs) {\textbf{Claim-pair}\\construction\\{\scriptsize cited, context, citing}};
\node[stage, right=0.78cm of pairs] (relations) {\textbf{Relation}\\prediction\\{\scriptsize GPT-4.1 few-shot}};
\node[stage, right=0.78cm of relations] (graph) {\textbf{\datasetauto{}}\\{\scriptsize canonical typed}\\{\scriptsize claim graph}};

\draw[flowarrow] (papers.east) -- (claims.west);
\draw[flowarrow] (papers.east) -- (contexts.west);
\draw[flowarrow] (claims.east) -- (pairs.west);
\draw[flowarrow] (contexts.east) -- (pairs.west);
\draw[flowarrow] (pairs) -- (relations);
\draw[flowarrow] (relations) -- (graph);

\node[note, below=0.38cm of papers] (paperNote) {raw paper text\\and metadata};
\node[note, above=0.34cm of claims] (claimNote) {candidate\\claim nodes};
\node[note, below=0.34cm of contexts] (contextNote) {local citation\\evidence};
\node[note, below=0.38cm of pairs] (pairNote) {classification\\instances};
\node[note, below=0.38cm of relations] (relNote) {\textcolor{claimflowsupport}{\texttt{support}},
\textcolor{claimflowextend}{\texttt{extend}},\\
\textcolor{claimflowqualify}{\texttt{qualify}},
\textcolor{claimflowrefute}{\texttt{refute}},\\
\textcolor{claimflowbackground}{\texttt{background}}};
\node[note, below=0.38cm of graph] (graphNote) {near-duplicate\\claims merged};

\draw[claimflowbackground!35, line width=0.35pt] (papers) -- (paperNote);
\draw[claimflowbackground!35, line width=0.35pt] (claims) -- (claimNote);
\draw[claimflowbackground!35, line width=0.35pt] (contexts) -- (contextNote);
\draw[claimflowbackground!35, line width=0.35pt] (pairs) -- (pairNote);
\draw[claimflowbackground!35, line width=0.35pt] (relations) -- (relNote);
\draw[claimflowbackground!35, line width=0.35pt] (graph) -- (graphNote);

\node[circle, fill=claimflowbackground, minimum size=5.5pt, inner sep=0pt, below=1.55cm of graph, xshift=-0.95cm] (g1) {};
\node[circle, fill=claimflowbackground, minimum size=5.5pt, inner sep=0pt, right=0.55cm of g1, yshift=0.33cm] (g2) {};
\node[circle, fill=claimflowbackground, minimum size=5.5pt, inner sep=0pt, right=0.58cm of g1, yshift=-0.35cm] (g3) {};
\node[circle, fill=claimflowbackground, minimum size=5.5pt, inner sep=0pt, right=1.12cm of g1, yshift=0.02cm] (g4) {};

\draw[relationedge, claimflowsupport] (g1) -- (g2);
\draw[relationedge, claimflowextend] (g1) -- (g3);
\draw[relationedge, claimflowqualify] (g2) -- (g4);
\draw[relationedge, claimflowrefute] (g3) -- (g4);
\node[font=\scriptsize, align=center, below=0.18cm of g3, xshift=0.35cm] {typed edges};
\end{tikzpicture}%
}
\caption{Pipeline for constructing \datasetauto{}.}
\label{fig:autograph_pipeline}
\end{figure*}

\subsection{Automatic Claim Identification}
\label{app:claim_identification}

We report the performance of the automatic claim identification model as used as an upstream component in constructing the large-scale claim graph. 

\paragraph{Experimental Setup.} We fine tune both SciBERT and DeBERTa on the expert-annotated \dataset{} using the same train-validation-test splits described in \S~\ref{sec:task}. Claim identification is formulated as a sentence-level binary classification task predicting whether a sentence expresses a scientific claim. Performance is evaluated using macro-precision, recall, and F1-score. 

\begin{table}[t]
\centering
\small
\setlength{\tabcolsep}{6pt}
\begin{tabular}{lccc}
\toprule
\textbf{Model} & \textbf{Precision} & \textbf{Recall} & \textbf{F1} \\
\midrule
SciBERT & 0.80 & 0.79 & 0.79 \\
\textbf{DeBERTa} & \textbf{0.83} & \textbf{0.82} & \textbf{0.82} \\
\bottomrule
\end{tabular}
\caption{Claim identification performance on the \dataset{} test set.}
\label{tab:claim_id_results}
\end{table}

\paragraph{Results.} Table~\ref{tab:claim_id_results} reports the results on the test split. DeBERTa achieves the strongest overall performance, outperforming SciBERT across all evaluation metrics. The strong performance of both models suggests that scientific claims can be identified reliably enough to support downstream construction of \datasetauto{}.

\subsection{Lightweight Canonicalization}
\label{app:canonicalization}

\begin{table}[t]
\centering
\begingroup
\definecolor{modelBest}{HTML}{357EDD}
\definecolor{modelGroupBg}{HTML}{F2F2F2}
\newcommand{\metricsetting}[1]{\begingroup\setlength{\fboxsep}{1.2pt}\colorbox{modelGroupBg}{\scriptsize\textsc{#1}}\endgroup}
\newcommand{\bestscore}[1]{\textcolor{modelBest}{\bf #1}}
\small
\setlength{\tabcolsep}{6pt}
\begin{tabular}{@{}l r@{}}
\toprule
{\bf Metric} & {\bf Value} \\
\midrule
\metricsetting{Quality} Cluster purity $\uparrow$ & {0.92} \\
\metricsetting{Risk} False merge rate $\downarrow$ & 0.08 \\
\metricsetting{Size} Average cluster size & 2.3 \\
\metricsetting{Compression} Reduction in claim nodes & {11\%} \\
\bottomrule
\end{tabular}
\endgroup
\caption{Evaluation of the lightweight canonicalization procedure.}
\label{tab:canonicalization_results}
\end{table}

Because automatically identified claim texts may contain repeated or near-duplicate surface realizations within the same paper, we apply a lightweight canonicalization procedure to reduce redundancy in the automatically inferred graph. 

\paragraph{Experimental Setup.} Each claim text is represented using a sentence-level embedding obtained from a DeBERTa encoder fine-tuned for claim identification. Within each paper, we compute pairwise cosine similarity between claim embeddings and greedily cluster claim texts whose similarity exceeds a threshold of $\tau=0.90$. For each cluster, we retain a single representative claim text corresponding to the longest surface realization, while edges incident on merged nodes are redirected to the retained representative. This procedure is intentionally conservative and merges only highly similar surface realizations. 

\paragraph{Diagnostic Analysis.} To evaluate whether canonicalization introduces substantial semantic distortion, we manually inspect randomly sampled $200$ canonicalization clusters and compare graph statistics before and after clustering. Table~\ref{tab:canonicalization_results} summarizes the results. We observe that canonicalization primarily merges near-duplicate claim surface forms while preserving the overall structural properties of the graph.

\subsection{Robustness Under Perturbation} 
\label{app:robustness}

\begin{figure}[t]
\centering
\scalebox{0.95}{
\begin{tikzpicture}
\begin{axis}[
    width=0.9\linewidth,
    height=5cm,
    xmin=10, xmax=30,
    ymin=0.75, ymax=1.0,
    xtick={10,20,30},
    ytick={0.8,0.85,0.9,0.95,1.0},
    xlabel={Perturbation Level (\%)},
    ylabel={Correlation},
    grid=major,
    grid style={draw=black!12, line width=0.4pt},
    thick,
    mark=*,
    mark size=2.5pt,
    enlargelimits=false,
    axis lines=left,
    axis line style={draw=black!35},
    tick align=outside,
]

\addplot+[claimflowbackground, line width=1.1pt, mark options={fill=claimflowbackground, draw=claimflowbackground}] 
coordinates {
    (10,0.96)
    (20,0.92)
    (30,0.89)
};

\end{axis}
\end{tikzpicture}
}
\caption{Correlation between original and perturbed aggregate graph statistics under increasing perturbation levels.}
\label{fig:robustness}
\end{figure}

We additionally evaluate whether downstream analyses derived from \datasetauto{} remain stable under realistic levels of automatic inference noise. 

\paragraph{Experimental Setup.} We simulate perturbations by randomly removing edges, randomly altering relation labels, and perturbing canonicalization clusters at varying noise levels. For each perturbed graph, we recompute the principal analyses presented in \S~\ref{sec:analysis}. 

\paragraph{Results.} Figure~\ref{fig:robustness} shows the correlation between original and perturbed graph statistics under increasing perturbation levels. We observe that the principal qualitative findings remain stable under moderate perturbations, suggesting that the aggregate conclusions derived from \datasetauto{} are robust to realistic pipeline noise.

\section{Extended Case Studies of Scientific Claim Evolution}
\label{app:case_studies}

\definecolor{cfSupport}{HTML}{19A974}
\definecolor{cfExtend}{HTML}{FFB700}
\definecolor{cfQualify}{HTML}{5E2CA5}
\definecolor{cfRefute}{HTML}{E7040F}
\definecolor{cfInk}{HTML}{222222}
\definecolor{cfPanel}{HTML}{F7F7F5}

\providecommand{\flowchip}[2]{%
  \begingroup\setlength{\fboxsep}{1.25pt}%
  \colorbox{#1}{\textcolor{white}{\tiny\bfseries #2}}%
  \endgroup%
}

\begin{figure*}[t]
\centering
\small
\usetikzlibrary{backgrounds}
\begin{tikzpicture}[
    x=1cm,
    y=1cm,
    source/.style={
        rounded corners=3pt,
        draw=cfInk!70,
        line width=.75pt,
        fill=white,
        align=left,
        text width=3.15cm,
        inner sep=4.5pt,
        font=\tiny
    },
    card/.style={
        rounded corners=3pt,
        line width=.75pt,
        fill=white,
        align=left,
        text width=3.05cm,
        inner sep=4.5pt,
        font=\tiny
    },
    supcard/.style={card, draw=cfSupport!85!black, fill=cfSupport!6},
    extcard/.style={card, draw=cfExtend!88!black, fill=cfExtend!12},
    qualcard/.style={card, draw=cfQualify!85!black, fill=cfQualify!6},
    refcard/.style={card, draw=cfRefute!85!black, fill=cfRefute!6},
    timeline/.style={-{Latex[length=1.6mm]}, line width=.9pt, draw=black!35},
    tick/.style={line width=.45pt, draw=black!35},
    branch/.style={line width=.9pt, rounded corners=4pt},
    support/.style={branch, -{Latex[length=2.2mm]}, draw=cfSupport},
    extend/.style={branch, -{Latex[length=2.2mm]}, draw=cfExtend!90!black},
    qualify/.style={branch, -{Latex[length=2.2mm]}, dashed, draw=cfQualify},
    refute/.style={branch, -{Latex[length=2.2mm]}, dashed, draw=cfRefute},
    title/.style={font=\bfseries, text=cfInk},
    year/.style={font=\tiny\bfseries, text=black!62, fill=cfPanel, inner xsep=1pt, inner ysep=.35pt},
    axislabel/.style={font=\tiny\bfseries, text=black!58}
]

\begin{scope}[on background layer]
\fill[cfPanel] (-.35,-.10) rectangle (14.65,5.25);
\fill[cfPanel] (-.35,-6.00) rectangle (14.65,-.85);
\end{scope}


\node[title, anchor=west] at (-.10,5.04) {(a) BERT evolution};

\node[source] (bert) at (1.85,2.65)
{\textbf{Source paper}\\[-1pt]
{\tiny\itshape BERT: Pre-training of Deep Bidirectional Transformers for Language Understanding}\\[3pt]
\textbf{Seed claim.}
BERT achieves state-of-the-art performance across NLP tasks without heavily engineered task-specific architectures.};

\draw[timeline] (3.95,2.65) -- (14.20,2.65);
\foreach \x/\yr in {4.20/2018,6.05/2020,10.10/2021,12.70/2023}{
    \draw[tick] (\x,2.59) -- (\x,2.71);
    \node[year] at (\x,2.65) {\yr};
}
\node[axislabel, anchor=south west] at (7.60,2.60) {publication year};
\draw[tick] (bert.east) -- (4.20,2.65);

\node[supcard] (bsup) at (6.05,4.02)
{\flowchip{cfSupport}{support} \hfill {\tiny\bfseries 2020}\\[2pt]
{\tiny\itshape BERT Knows Punta Cana is not just beautiful, it's gorgeous}\\[2pt]
\textbf{Claim.} BERT encodes scalar-adjective meaning and ranks intensity better than static embeddings.};

\node[extcard] (bext) at (6.05,1.22)
{\flowchip{cfExtend!90!black}{extend} \hfill {\tiny\bfseries 2020}\\[2pt]
{\tiny\itshape CodeBERT: A Pre-Trained Model for Programming and Natural Languages}\\[2pt]
\textbf{Claim.} A hybrid objective improves CodeBERT representations for paired code and language.};

\node[qualcard] (bqual) at (10.10,4.02)
{\flowchip{cfQualify}{qualify} \hfill {\tiny\bfseries 2021}\\[2pt]
{\tiny\itshape Understanding by Understanding Not}\\[2pt]
\textbf{Claim.} Pretrained models such as BERT fail to correctly handle negation in NLU.};

\node[refcard] (bref) at (12.70,1.22)
{\flowchip{cfRefute}{refute} \hfill {\tiny\bfseries 2023}\\[2pt]
{\tiny\itshape A Universal Discriminator for Zero-Shot Generalization}\\[2pt]
\textbf{Claim.} A universal discriminator achieves state-of-the-art zero-shot results on T0.};

\draw[support] (6.05,2.65) -- (bsup.south);
\draw[extend] (6.05,2.65) -- (bext.north);
\draw[qualify] (10.10,2.65) -- (bqual.south);
\draw[refute] (12.70,2.65) -- (bref.north);


\node[title, anchor=west] at (-.10,-1.18) {(b) BLEU evolution};

\node[source] (bleu) at (1.85,-3.45)
{\textbf{Source paper}\\[-1pt]
{\tiny\itshape Bleu: a Method for Automatic Evaluation of Machine Translation}\\[3pt]
\textbf{Seed claim.}
BLEU is an automatic machine-translation metric that correlates highly with human evaluation.};

\draw[timeline] (3.95,-3.45) -- (14.20,-3.45);
\foreach \x/\yr in {4.20/2002,6.00/2014,7.10/2016,11.45/2020,12.70/2021}{
    \draw[tick] (\x,-3.51) -- (\x,-3.39);
    \node[year] at (\x,-3.45) {\yr};
}
\node[axislabel, anchor=south west] at (8.50,-3.50) {publication year};
\draw[tick] (bleu.east) -- (4.20,-3.45);

\node[supcard] (lsup) at (6.00,-2.10)
{\flowchip{cfSupport}{support} \hfill {\tiny\bfseries 2014}\\[2pt]
{\tiny\itshape Testing for Significance of Increased Correlation with Human Judgment}\\[2pt]
\textbf{Claim.} Many new metrics do not significantly outperform BLEU under human-correlation tests.};

\node[extcard] (lext) at (7.10,-4.80)
{\flowchip{cfExtend!90!black}{extend} \hfill {\tiny\bfseries 2016}\\[2pt]
{\tiny\itshape Machine Translation Evaluation Meets Community Question Answering}\\[2pt]
\textbf{Claim.} TER, METEOR, and BLEU help community question-answering evaluation.};

\node[refcard] (lref) at (11.45,-4.80)
{\flowchip{cfRefute}{refute} \hfill {\tiny\bfseries 2020}\\[2pt]
{\tiny\itshape On The Evaluation of Machine Translation Systems Trained With Back-Translation}\\[2pt]
\textbf{Claim.} BLEU cannot capture human preferences when references are translationese.};

\node[qualcard] (lqual) at (12.70,-2.10)
{\flowchip{cfQualify}{qualify} \hfill {\tiny\bfseries 2021}\\[2pt]
{\tiny\itshape As Easy as 1, 2, 3}\\[2pt]
\textbf{Claim.} BLEU may fail to adequately flag numerical translation errors.};

\draw[support] (6.00,-3.45) -- (lsup.south);
\draw[extend] (7.10,-3.45) -- (lext.north);
\draw[refute] (11.45,-3.45) -- (lref.north);
\draw[qualify] (12.70,-3.45) -- (lqual.south);

\end{tikzpicture}

\caption{
Timeline-centered claim evolution for two influential NLP papers.
Each panel roots the original seed claim at the start of the publication timeline and places later interacting claims as branches at their publication years.
The branch color and line style encode the ClaimFlow relation type: support, extend, qualify, or refute.
}
\label{fig:case_study_trajectories}
\end{figure*}

We provide additional qualitative analyses of representative claim trajectories extracted for \datasetauto{}. These case studies complement the aggregate analyses presented in \S~\ref{sec:analysis} by illustrating how influential claims evolve through \texttt{support}, \texttt{extension}, \texttt{qualification}, and \texttt{refutation} over time. Representative downstream examples are shown in Tables~\ref{tab:bert_case_examples} and~\ref{tab:bleu_case_examples}.

\providecommand{\definecolor}[3]{}
\providecommand{\textcolor}[2]{#2}
\providecommand{\colorbox}[2]{#2}
\definecolor{cfBackgroundBg}{HTML}{EBF3FD}
\definecolor{cfBackgroundFg}{HTML}{357EDD}
\definecolor{cfSupportBg}{HTML}{E8F7F1}
\definecolor{cfSupportFg}{HTML}{19A974}
\definecolor{cfExtendBg}{HTML}{FFF4CC}
\definecolor{cfExtendFg}{HTML}{FFB700}
\definecolor{cfQualifyBg}{HTML}{EFEAF6}
\definecolor{cfQualifyFg}{HTML}{5E2CA5}
\definecolor{cfRefuteBg}{HTML}{FDE6E7}
\definecolor{cfRefuteFg}{HTML}{E7040F}
\newcommand{\cfrel}[3]{\begingroup\setlength{\fboxsep}{1.3pt}\colorbox{#1}{\textcolor{#2}{\scriptsize\textsc{#3}}}\endgroup}
\newcommand{\cfbackground}{\cfrel{cfBackgroundBg}{cfBackgroundFg}{background}}
\newcommand{\cfsupport}{\cfrel{cfSupportBg}{cfSupportFg}{support}}
\newcommand{\cfextend}{\cfrel{cfExtendBg}{cfExtendFg}{extend}}
\newcommand{\cfqualify}{\cfrel{cfQualifyBg}{cfQualifyFg}{qualify}}
\newcommand{\cfrefute}{\cfrel{cfRefuteBg}{cfRefuteFg}{refute}}


\begin{table*}[t]
\centering
\small
\setlength{\tabcolsep}{4pt}
\renewcommand{\arraystretch}{1.12}
\begin{tabularx}{\textwidth}{>{\raggedright\arraybackslash}p{0.06\textwidth} >{\raggedright\arraybackslash}p{0.10\textwidth} >{\raggedright\arraybackslash}p{0.39\textwidth} >{\raggedright\arraybackslash}X}
\toprule
\textbf{Year} & \textbf{Relation} & \textbf{Downstream claim} & \textbf{Interpretation} \\
\midrule
\multicolumn{4}{p{0.96\textwidth}}{\textbf{Seed claim:} BERT achieves strong performance across NLP tasks without heavily engineered task-specific architectures.} \\
\midrule

2020 & \cfsupport &
\textbf{Scalar adjective semantics.}
\emph{``BERT encodes rich knowledge about the semantics of scalar adjectives and provides better quality intensity rankings than static embeddings and previous models.''}
\newline {\scriptsize BERT Knows Punta Cana is not just beautiful, it's gorgeous}
&
Gives a concrete linguistic case where bidirectional contextual representations outperform non-contextual baselines, supporting the representational part of BERT's claim. \\

2020 & \cfextend &
\textbf{Code and natural language.}
\emph{``Training CodeBERT with a hybrid objective function improves its performance.''}
\newline {\scriptsize CodeBERT: A Pre-Trained Model for Programming and Natural Languages}
&
Extends BERT-style masked pretraining from ordinary text into paired programming-language and natural-language representations. \\


2021 & \cfqualify &
\textbf{Negation in NLU.}
\emph{``Current pre-trained language models like BERT fail to correctly handle negation in language understanding tasks.''}
\newline {\scriptsize Understanding by Understanding Not: Modeling Negation in Language Models}
&
Makes the limitation explicit: high benchmark performance can coexist with brittle handling of logically decisive words such as negation. \\



2023 & \cfrefute &
\textbf{Zero-shot universal discrimination.}
\emph{``The universal discriminator (UD) achieves state-of-the-art zero-shot results on the T0 benchmark.''}
\newline {\scriptsize A Universal Discriminator for Zero-Shot Generalization}
&
Gives an explicit counterexample to BERT-style adaptation as the default path to broad task generalization: a universal discriminator can win in a zero-shot benchmark setting. \\

\bottomrule
\end{tabularx}
\caption{
Representative downstream interactions for the BERT case study.
The examples illustrate the life cycle of a high-impact model claim: broad support, transfer into new modalities and domains, and later qualifications about reasoning, structure, and zero-shot alternatives.
}
\label{tab:bert_case_examples}
\end{table*}


\begin{table*}[t]
\centering
\small
\setlength{\tabcolsep}{4pt}
\renewcommand{\arraystretch}{1.12}
\begin{tabularx}{\textwidth}{>{\raggedright\arraybackslash}p{0.06\textwidth} >{\raggedright\arraybackslash}p{0.10\textwidth} >{\raggedright\arraybackslash}p{0.39\textwidth} >{\raggedright\arraybackslash}X}
\toprule
\textbf{Year} & \textbf{Relation} & \textbf{Downstream claim} & \textbf{Interpretation} \\
\midrule
\multicolumn{4}{p{0.96\textwidth}}{\textbf{Seed Claim:} BLEU provides an automatic MT metric that correlates highly with human evaluation.} \\
\midrule

2014 & \cfsupport &
\textbf{Metric significance testing.}
\emph{``A high proportion of new metrics do not significantly outperform BLEU when evaluated with the introduced significance test.''}
\newline {\scriptsize Testing for Significance of Increased Correlation with Human Judgment}
&
Supports BLEU's durability: even later metrics often fail to beat it significantly when correlation with human judgment is tested rigorously. \\

2016 & \cfextend &
\textbf{Community question answering.}
\emph{``Using machine translation evaluation measures such as TER, METEOR, and BLEU can enhance the performance of the cQA task.''}
\newline {\scriptsize Machine Translation Evaluation Meets Community Question Answering}
&
Extends BLEU from MT evaluation into answer selection, where translation-style overlap becomes a feature for a different NLP problem. \\

2021 & \cfqualify &
\textbf{Numerical translation errors.}
\emph{``Existing de facto standard metrics such as BLEU may fail to adequately flag numerical translation errors.''}
\newline {\scriptsize As Easy as 1, 2, 3: Behavioral Testing of NMT Systems for Numerical Translation}
&
Exposes a high-stakes blind spot: a single wrong number may preserve much of the n-gram overlap while changing the meaning completely. \\


2020 & \cfrefute &
\textbf{Translationese references.}
\emph{``BLEU cannot capture human preferences because references are translationese when source sentences are natural text.''}
\newline {\scriptsize On The Evaluation of Machine Translation Systems Trained With Back-Translation}
&
Directly refutes the proxy claim in a concrete MT setting: reference artifacts can make BLEU reward outputs that humans do not prefer. \\

\bottomrule
\end{tabularx}
\caption{
Representative downstream interactions for the BLEU case study.
The examples show BLEU's transition from a durable automatic proxy to a cautionary baseline whose limits become visible in numeracy, generation, translationese, and human-preference settings.
}
\label{tab:bleu_case_examples}
\end{table*}

\subsection{BERT: Evolution of Bidirectional Pretraining Claims}

We first analyze the evolution of the claim introduced by \citet{devlin-etal-2019-bert} that deep bidirectional pretraining substantially improves downstream NLP performance. Figure~\ref{fig:case_study_trajectories}(a) visualizes a representative subset of downstream claim interactions associated with this trajectory, while Table~\ref{tab:bert_case_examples} presents concrete examples extracted from \datasetauto{}.

In the years immediately following its introduction, the claim received a large number of support interactions from papers reporting strong empirical gains across diverse NLP tasks. Early downstream work further extends the original claim into new domains and modalities, including biomedical NLP, multilingual transfer, and multi-modal representation learning~\citep{NEURIPS2019_c74d97b0, kanakarajan-etal-2019-saama, conneau-etal-2020-unsupervised}. These interactions suggest that the claim rapidly became a convergence point for subsequent research, attracting engagement from a broad range of sub-fields within NLP~\citep{liu2019roberta, clark2020electric}. 

As the trajectory evolves, later work increasingly introduces qualification interactions that constrain the scope of the original claim. Several downstream claims question whether pretrained bidirectional representations alone are sufficient for reasoning-intensive tasks or efficient adaptation under limited resources~\citep{ribeiro-etal-2020-beyond, kassner-schutze-2020-negated}. Other work shifts attention toward prompting-based and zero-shot paradigms~\citep{brown2020language, sanh2022multitask}, partially challenging the assumption that task-specific fine-tuning of pretrained representations is the dominant route to generalization. For example, the zero-shot universal discriminator trajectory shown in Table~\ref{tab:bert_case_examples} provides an explicit counterexample to BERT-style adaptation as the default mechanism for broad task transfer. 

Importantly, the trajectory contains relatively little sustained adversarial rejection of the original claim itself. Instead, the dominant pattern is progressive refinement, where subsequent work extends, constrains, or contextualizes the original claim while preserving its broader influence on NLP research. This pattern closely aligns with the aggregate analyses in \S~\ref{sec:analysis}, which show that influential claims in NLP more often evolve through extension and qualification than through direct contradiction.

\subsection{BLEU: Evolution of Evaluation Claims}

We next examine the evolution of the claim introduced by \citet{papineni-etal-2002-bleu} that BLEU provides an effective automatic proxy for machine translation quality. Figure~\ref{fig:case_study_trajectories}(b) visualizes representative downstream interactions associated with this claim trajectory, while Table~\ref{tab:bleu_case_examples} provides concrete examples from \datasetauto{}.

Early interactions overwhelmingly support or reuse the original claim, reflecting the rapid adoption of BLEU as a standard evaluation metric in machine translation research~\citep{koehn-2004-statistical, callison-burch-etal-2007-meta}. During this period, BLEU functions not merely as a metric but as a shared evaluation infrastructure, enabling reproducible comparison across systems and benchmarks~\citep{post-2018-call}. The support example shown in Table~\ref{tab:bleu_case_examples} further illustrates the metrics' durability: even later evaluation metrics often fail to significantly outperform BLEU under rigorous statistical testing.

Over time, however, later work increasingly qualifies and occasionally refutes the original claim. Several downstream claims identify important blind spots in BLEU's ability to capture semantic adequacy, numerical correctness, and human preferences~\citep{novikova-etal-2017-need, sellam-etal-2020-bleurt}. The numerical translation example shown in Table~\ref{tab:bleu_case_examples} demonstrates how high lexical overlap can preserve BLEU scores even when the meaning of a translation changes substantially. Other work directly challenges the metrics' validity in realistic translation settings by arguing that translationese artifacts can cause BLEU to reward outputs that humans do not prefer~\citep{li-etal-2025-evaluating-wmt}. 

At the same time, the trajectory also contains extension interactions that adapt BLEU-style overlap metrics to tasks outside machine translation, such as community question answering. This illustrates how influential methodological claims may continue to propagate into new application domains even as their original assumptions become increasingly constrained.


\section{Supplementary Results}
\label{app:add_results}

\subsection{Structural Evolution of the Claim Graph}
\label{subsec:structural_evolution_app}


For each year $t$, we construct a cumulative directed graph \(G_t = (V_t, E_t)\), where $V_t$ is the set of claims introduced up to year $t$, and $E_t$ contains directed edges $(h_{citing} \rightarrow h_{cited})$, corresponding to claim-claim relations up to $t$. To assess whether the claim graph densifies over time, we compute edge density \(\delta(G_t) = \frac{|E_t|}{|V_t|(|V_t|-1)}\), which measures the fraction of realized claim-claim interactions among all possible directed pairs. Intuitively, higher density indicates that claims increasingly connect to many other claims, reflecting a more tightly interlinked body of work. To assess whether the graph stratifies, we measure modularity $Q(G_t)$, which quantifies the extent to which claims form groups with dense internal connections but relatively sparse connections to other groups. Intuitively, higher modularity indicates stronger clustering of claims into distinct thematic or methodological substructures.

\begin{figure}[t]
    \centering
    \begin{subfigure}{\linewidth}
        \centering
        \includegraphics[width=0.82\linewidth]{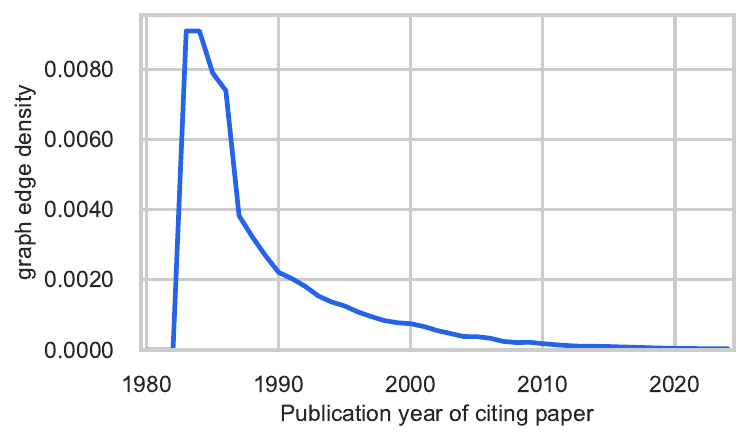}
        \caption{Densification of the claim--claim graph over time.}
        \label{fig:rq5_densification_app}
    \end{subfigure}

        \begin{subfigure}{\linewidth}
        \centering
        \includegraphics[width=0.82\linewidth]{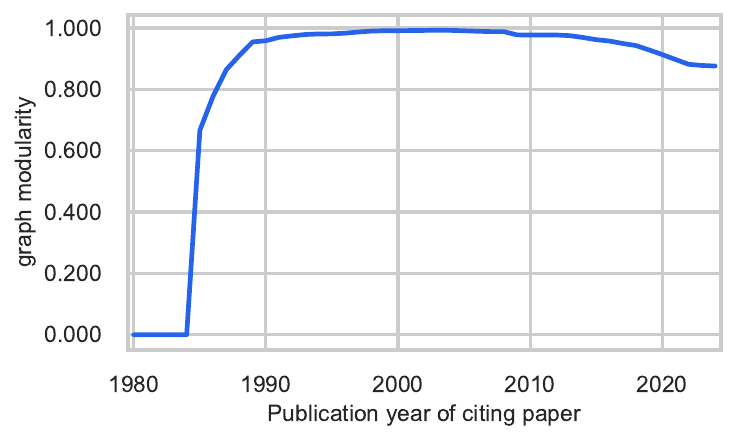}
        \caption{Increasing stratification of the claim--claim interaction graph over time.}
        \label{fig:rq5_modularity_app}
    \end{subfigure}
    \caption[Separate views of the density and modularity trends.]{Separate views of the density and modularity trends summarized in Figure~\ref{fig:rq5_structural_evolution}.}
    \label{fig:rq5_structural_evolution_app}
\end{figure}

Figure~\ref{fig:rq5_structural_evolution_app} separates the two trends summarized in the main paper. The density curve remains extremely low overall after the initial small-graph regime, indicating that new claims connect to only a limited subset of prior claims rather than forming a broadly interconnected graph. In contrast, modularity increases over time before plateauing and declining in recent years, suggesting that the claim graph becomes more stratified as the field specializes and later partially reconnects through shared modeling paradigms.

As the field expanded in the 1990s and early 2000s, the emergence of large resources and standardized evaluations, such as treebanks~\citep{marcus-etal-1993-building}, lexical databases~\citep{miller1995wordnet}, and task-specific benchmarks, encouraged specialization, leading claims to cluster around distinct tasks, representations, and evaluation regimes. The subsequent plateau and recent decline in modularity are consistent with methodological convergence in modern NLP. Since the mid-2010s, contextualized and pretrained models~\citep{devlin-etal-2019-bert} have been widely adopted across diverse tasks, increasing cross-subfield engagement and weakening previously isolated clusters. This trend has been further reinforced by multi-task and unified benchmarks~\citep{wang-etal-2018-glue, wang2019superglue}, which explicitly promote cross-task comparison, and by the rise of foundation models~\citep{bommasani2021opportunities}, which frame many new claims as analyses or extensions of shared model families rather than task-specific innovations.

\begin{table}[!ht]
    \centering
\begingroup
\definecolor{modelBest}{HTML}{357EDD}
\definecolor{modelGroupBg}{HTML}{F2F2F2}
\newcommand{\modelsetting}[1]{\begingroup\setlength{\fboxsep}{1.2pt}\colorbox{modelGroupBg}{\scriptsize\textsc{#1}}\endgroup}
\newcommand{\bestscore}[1]{\textcolor{modelBest}{\bf #1}}
    \small
    \setlength{\tabcolsep}{5pt}
    \begin{tabular}{@{}l l c@{}}
    \toprule 
    {\bf Input} & {\bf Model} & {\bf Macro-F1} \\
    \midrule
    \multirow{4}{*}{\modelsetting{Context} ($s$)} & BERT & 0.44 \\
     & RoBERTa & 0.46 \\
     & SciBERT & 0.51 \\
     & \textbf{DeBERTa} & \textbf{0.54} \\

     \midrule

     \multirow{4}{*}{\modelsetting{Cited + context} ($h_{cited}, s$)} & BERT & 0.45 \\
     & RoBERTa & 0.50 \\
     & SciBERT & 0.56 \\
     & \textbf{DeBERTa} & \textbf{0.60} \\

     \midrule

     \multirow{4}{*}{\modelsetting{Full} ($h_{cited}, s, h_{citing}$)} & BERT & 0.50 \\
     & RoBERTa & 0.70 \\
     & SciBERT & 0.75 \\
     & \textbf{DeBERTa} & \textbf{0.76} \\

     \bottomrule
    \end{tabular}
\endgroup
    \caption{Effect of different input components on claim relation classification performance.}
    \label{tab:plm_abl}
\end{table}

\begin{table}[!ht]
    \centering
    \small
    \setlength{\tabcolsep}{5pt}
    \begin{tabular}{lccc}
    \toprule
    \textbf{Model} & \textbf{0-shot} & \textbf{1-shot} & \textbf{4-shot} \\
    \midrule
    GPT-3.5-Turbo & 0.60 & 0.65 &  0.75\\
    GPT-4o & 0.61 & 0.70 & 0.78 \\
    {\bf GPT-4.1} & {\bf 0.63} & {\bf 0.73} & {\bf 0.80} \\
    LLaMA-3-70B & 0.58 & 0.61 & 0.70 \\
    Mixtral-8x70B & 0.60 & 0.65 & 0.73 \\
    \bottomrule
    \end{tabular}
    \caption{LLM performance (\texttt{macro-F1}) with different number of training examples.}
    \label{tab:llm_shots}
\end{table}

\begin{table*}[t]
\centering
\small
\setlength{\tabcolsep}{8pt}
\newcommand{\labelbest}[1]{\textcolor{claimflowbackground}{\textbf{#1}}}
\begin{tabular}{lccccc}
\toprule
\textbf{Model} & \textcolor{claimflowsupport}{\texttt{support}} & \textcolor{claimflowextend}{\texttt{extend}} & \textcolor{claimflowqualify}{\texttt{qualify}} & \textcolor{claimflowrefute}{\texttt{refute}} & \textcolor{claimflowbackground}{\texttt{background}} \\
\midrule
SciBERT         & 0.78 & 0.70 & 0.69 & 0.76 & 0.82 \\
DeBERTa         & 0.80 & 0.73 & 0.71 & 0.78 & 0.83 \\
\midrule
GPT-4o (few-shot) & 0.80 & 0.76 & 0.75 & 0.78 & 0.83 \\
\textbf{GPT-4.1 (few-shot)} & \textbf{0.82} & \textbf{0.78} & \textbf{0.77} & \textbf{0.80} & \textbf{0.85} \\
\bottomrule
\end{tabular}
\caption{Label-wise \texttt{macro-F1} scores for model performance on automatic claim-relation classification.}
\label{tab:labelwise_f1}

\end{table*}

\begin{table*}[!ht]
    \centering
    \small
    \setlength{\tabcolsep}{5pt}
    \newcommand{\relsupport}{\textcolor{claimflowsupport}{\texttt{support}}}
    \newcommand{\relextend}{\textcolor{claimflowextend}{\texttt{extend}}}
    \newcommand{\relqualify}{\textcolor{claimflowqualify}{\texttt{qualify}}}
    \newcommand{\relrefute}{\textcolor{claimflowrefute}{\texttt{refute}}}
    \newcommand{\relbackground}{\textcolor{claimflowbackground}{\texttt{background}}}
    \begin{adjustbox}{width=\textwidth, center}
    \begin{tabular}{>{\raggedright\arraybackslash}p{0.20\textwidth} >{\raggedright\arraybackslash}p{0.42\textwidth} >{\raggedright\arraybackslash}p{0.22\textwidth} p{0.08\textwidth}}
    \toprule
    \textbf{Error} & \textbf{Description} & \textbf{Typical confusion} & \textbf{Share} \\
    \midrule
    Extend--Qualify ambiguity & Difficulty distinguishing claim extension from scope restriction & \relextend{} $\leftrightarrow$ \relqualify{} & \textbf{$\sim$$30\%$} \\

    \addlinespace

    Implicit stance & Stance expressed through hedging or indirect comparison & \relsupport{} $\rightarrow$ \relbackground{} & $\sim$$20\%$ \\

    \addlinespace

    Evidence mismatch & Model aligns citing claim with the wrong cited claim & \relsupport{} $\rightarrow$ \relextend{} & $\sim$$15\%$ \\

    \addlinespace

    Weak refutation cues & Disagreement expressed without explicit negation & \relrefute{} $\rightarrow$ \relqualify{} & $\sim$$15\%$ \\

    \addlinespace

    Context under-specification & Citation sentence lacks sufficient detail & \textcolor{black!55}{any} $\rightarrow$ \relbackground{} & $\sim$$10\%$ \\

    \addlinespace

    Other & Miscellaneous rare cases & -- & $\sim$$10\%$ \\

    \bottomrule
    \end{tabular}
    \end{adjustbox}
    \caption{Common error types observed in automatic claim relation classification.}
    \label{tab:error_qual}
\end{table*}

\end{document}